\documentclass[11pt]{article} % Stat tech report

\newcommand{\doctype}{TECH}

\usepackage{amsfonts,graphicx,subfigure}
\usepackage{amsfonts,amsmath,amssymb,amsthm}
\usepackage{epic,eepic,eepicemu}
\usepackage{epsf}
\usepackage{epsfig}
\usepackage{fancyheadings}
\usepackage{graphics}
\usepackage{psfrag,latexsym}

\usepackage{ifthen}

\ifthenelse{\equal{\doctype}{MP}}
{ 
\typeout{---------> Minor modifications}
\typeout{}
\oddsidemargin .00in
\evensidemargin .00in
\marginparwidth 0.15 true in
\addtolength{\headsep}{0.25in}
\textheight 8.1 true in
\textwidth 6.0 true in
\renewcommand{\baselinestretch}{1.9}
}
{ 
\typeout{------> Setting Stat. tech report formatting}
\typeout{}

\setlength{\topmargin}{0in}
\setlength{\headheight}{0in}
\setlength{\headsep}{0in}
\setlength{\textheight}{8.5 in}
\setlength{\oddsidemargin}{0in}
\setlength{\evensidemargin}{0in}
\setlength{\textwidth}{6.5in}
\addtolength{\headsep}{0.25in}
\marginparwidth 0.07 true in

}

\newenvironment{carlist}
 {\begin{list}{$\bullet$}
 {\setlength{\topsep}{0in} \setlength{\partopsep}{0in}
  \setlength{\parsep}{0in} \setlength{\itemsep}{\parskip}
  \setlength{\leftmargin}{0.07in} \setlength{\rightmargin}{0.08in}
  \setlength{\listparindent}{0in} \setlength{\labelwidth}{0.08in}
  \setlength{\labelsep}{0.1in} \setlength{\itemindent}{0in}}}
 {\end{list}}

\newcommand{\bcar}{\begin{carlist}}
\newcommand{\ecar}{\end{carlist}}

\newenvironment{carliste}
 {\begin{list}x
 {\setlength{\topsep}{0in} \setlength{\partopsep}{0in}
  \setlength{\parsep}{0in} \setlength{\itemsep}{\parskip}
  \setlength{\leftmargin}{0.07in} \setlength{\rightmargin}{0.08in}
  \setlength{\listparindent}{0in} \setlength{\labelwidth}{0.08in}
  \setlength{\labelsep}{0.1in} \setlength{\itemindent}{0in}}}
 {\end{list}}

\newcommand{\bcare}{\begin{carliste}}
\newcommand{\ecare}{\end{carliste}}

\newcommand{\dom}{\ensuremath{\operatorname{dom}}}

\newcommand{\Locset}{\ensuremath{\operatorname{LOCAL}}}

\newcommand{\treegr}{\ensuremath{T}}

\newcommand{\meanpar}{\ensuremath{\mu}}

\newcommand{\taupar}{\ensuremath{\tau}}

\makeatletter
\long\def\@makecaption#1#2{
        \vskip 0.8ex
        \setbox\@tempboxa\hbox{\small {\bf #1:} #2}
        \parindent 1.5em  %% How can we use the global value of this???
        \dimen0=\hsize
        \advance\dimen0 by -3em
        \ifdim \wd\@tempboxa >\dimen0
                \hbox to \hsize{
                        \parindent 0em
                        \hfil 
                        \parbox{\dimen0}{\def\baselinestretch{0.96}\small
                                {\bf #1.} #2
                                } 
                        \hfil}
        \else \hbox to \hsize{\hfil \box\@tempboxa \hfil}
        \fi
        }
\makeatother

\long\def\comment#1{}

\makeatletter
\def\@cite#1#2{[\if@tempswa #2 \fi #1]}
\makeatother

\makeatletter
\long\def\barenote#1{
    \insert\footins{\footnotesize
    \interlinepenalty\interfootnotelinepenalty 
    \splittopskip\footnotesep
    \splitmaxdepth \dp\strutbox \floatingpenalty \@MM
    \hsize\columnwidth \@parboxrestore
    {\rule{\z@}{\footnotesep}\ignorespaces
      #1\strut}}}
\makeatother
\newcommand{\Partc}{\ensuremath{A}}
\newcommand{\Dualc}{\ensuremath{\Partc^*}}

\newcommand{\Ind}{\ensuremath{\mathbb{I\,}}}

\newcommand{\Margset}{\ensuremath{\operatorname{MARG}}}

\newcommand{\bit}{\begin{itemize}}
\newcommand{\eit}{\end{itemize}}
\newcommand{\ben}{\begin{enumerate}}
\newcommand{\een}{\end{enumerate}}

\newcommand{\bear}{\begin{eqnarray}}
\newcommand{\eear}{\end{eqnarray}}

\newcommand{\vtiny}{\vspace*{.1in}}

\newcommand{\fn}{\footnotesize}
\newcommand{\uni}{\ensuremath{\mathcal{U}}}

\newcommand{\clipotvec}{\ensuremath{{\boldsymbol{\clipot}}}}

\newcommand{\Info}{\ensuremath{I}}
\newcommand{\Ent}{\ensuremath{H}}

\newcommand{\tract}{\mathfrak{T}}

\newcommand{\df}{\ensuremath{d}}

\newcommand{\eparams}{{\ensuremath{\eparam^*}}}

\newcommand{\sumind}{\ensuremath{\alpha}}

\newcommand{\statenum}{{\ensuremath{m}}}
\newcommand{\nodenum}{\ensuremath{N}}

\newcommand{\neigh}{\ensuremath{\Gamma}}

\newcommand{\clipot}{\ensuremath{\phi}}

\newcommand{\eparam}{\ensuremath{\theta}}

\newcommand{\statesp}{\ensuremath{{\scr{X}}}}

\newcommand{\statespn}{\ensuremath{\statesp^{\nodenum}}}

\newcommand{\order}{{\mathcal{O}}}

\newcommand{\inprod}[2]{\ensuremath{\langle #1 , \, #2 \rangle}}

\newcommand{\covv}[3]{\ensuremath{\cov_{#3}\{#1, \, #2\}}}

\newcommand{\graph}{\ensuremath{G}}

\newcommand{\vertex}{\ensuremath{V}}
\newcommand{\edge}{\ensuremath{E}}

\newcommand{\pdif}[2]{\ensuremath{\frac{\partial{#1}}{\partial{#2}}}}
\newcommand{\pdifftwo}[3]{\ensuremath{\frac{\partial^2{#1}}{\partial{#2}
\, \partial{#3}}}}

\newcommand{\Exs}{\ensuremath{{\mathbb{E}}}}

\newcommand{\beq}{\begin{quotation}}
\newcommand{\enq}{\end{quotation}}

\newcommand{\estart}{\begin{equation}}
\newcommand{\eend}{\end{equation}}

\newcommand{\widgraph}[2]{\includegraphics[keepaspectratio,width=#1]{#2}}

\newcommand{\scr}[1]{\ensuremath{\mathcal{#1}}}

\newcommand{\defn}{\ensuremath{:  =}}

\newcommand{\Ysca}{{{Y}}}

\newcommand{\bec}{\begin{center}}
\newcommand{\enc}{\end{center}}

\newcommand{\beit}{\begin{itemize}}
\newcommand{\enit}{\end{itemize}}

\newcommand{\been}{\begin{enumerate}}
\newcommand{\enen}{\end{enumerate}}

\newcommand{\comsl}{\begin{slide}}
\newcommand{\comspor}{\begin{slide*}}

\newcommand{\comsld}[2]{\begin{slide}[#1,#2]}
\newcommand{\comspord}[2]{\begin{slide*}[#1,#2]}

\newcommand{\mendsl}{\end{slide}}
\newcommand{\mendspo}{\end{slide*}}

\newcommand{\estim}[1]{\ensuremath{\widehat{#1}}}

\newcommand{\real}{\ensuremath{{\mathbb{R}}}}

\DeclareMathOperator{\var}{var}

\DeclareMathOperator{\cov}{cov}

\DeclareMathOperator{\trace}{trace}

\theoremstyle{plain}

\newtheorem{theo}{Theorem}[section]

\newtheorem{lem}{Lemma}[section]
\newtheorem{prop}{Proposition}[section]
\newtheorem{cor}{Corollary}[section]

\theoremstyle{definition} 

\newtheorem{nota}{Notation}[section]
\newtheorem{de}{Definition}[section]
\newtheorem{exa}{Example}[section]
\newtheorem{as}{Assumption}[section]
\newtheorem{alg}{Algorithm}[section]

\newcommand{\btheo}{\begin{theo}}
\newcommand{\bde}{\begin{de}}
\newcommand{\ble}{\begin{lem}}
\newcommand{\bpr}{\begin{prop}}
\newcommand{\bno}{\begin{nota}}
\newcommand{\bex}{\begin{exa}}
\newcommand{\bcor}{\begin{cor}}
\newcommand{\spro}{\begin{proof}}
\newcommand{\bas}{\begin{as}}
\newcommand{\balg}{\begin{alg}}

\newcommand{\etheo}{\end{theo}}
\newcommand{\ede}{\end{de}}
\newcommand{\ele}{\end{lem}}
\newcommand{\epr}{\end{prop}}
\newcommand{\eno}{\end{nota}}
\newcommand{\eex}{\end{exa}}
\newcommand{\ecor}{\end{cor}}
\newcommand{\fpro}{\end{proof}}
\newcommand{\eas}{\end{as}}
\newcommand{\ealg}{\end{alg}}

\theoremstyle{plain}

\newtheorem{theos}{Theorem}
\newtheorem{props}{Proposition}
\newtheorem{lems}{Lemma}
\newtheorem{cors}{Corollary}

\theoremstyle{definition}
\newtheorem{exas}{Example}
\newtheorem{algs}{Algorithm}
\newtheorem{asss}{Asumption}
\newtheorem{defns}{Definition}

\newcommand{\btheos}{\begin{theos}}
\newcommand{\etheos}{\end{theos}}
\newcommand{\bprops}{\begin{props}}
\newcommand{\eprops}{\end{props}}
\newcommand{\bdes}{\begin{defns}}
\newcommand{\edes}{\end{defns}}
\newcommand{\blems}{\begin{lems}}
\newcommand{\elems}{\end{lems}}
\newcommand{\bcors}{\begin{cors}}
\newcommand{\ecors}{\end{cors}}
\newcommand{\bexs}{\begin{exas}}
\newcommand{\eexs}{\end{exas}}
\newcommand{\balgs}{\begin{algs}}
\newcommand{\ealgs}{\end{algs}}
\newcommand{\bass}{\begin{asss}}
\newcommand{\eass}{\end{asss}}

\newcommand{\Parta}{\ensuremath{A}}

\newcommand{\Sur}{\ensuremath{B}}
\newcommand{\Dsur}{\ensuremath{\Sur^*}}

\newcommand{\dnum}{\ensuremath{n}}

\newcommand{\vnum}{\ensuremath{N}}

\newcommand{\edgew}{\ensuremath{\rho}}
\newcommand{\edgewvec}{\ensuremath{\mathbf{\rho}}}

\newcommand{\alsnr}{\ensuremath{\alpha}}

\newcommand{\epy}{\ensuremath{\gamma}}

\newcommand{\pd}[2]{\ensuremath{p(#2 \, ; \, #1)}}

\newcommand{\zmuvecs}[2]{\ensuremath{\estim{z}^{\,#1}(#2; \mu)}}

\newcommand{\tauvec}{\ensuremath{{\tau}}}

\newcommand{\myvec}{\ensuremath{\omega}}
\newcommand{\Lcon}{\ensuremath{L}}

\newcommand{\Ccon}{\ensuremath{C}}

\newcommand{\Cbethe}{\Dsur_{\edgew}}
\newcommand{\Cbsur}{\Sur_{\edgew}}

\newcommand{\tauparr}[1]{\ensuremath{\taupar(#1)}}

\newcommand{\Projt}{\ensuremath{\Pi^T}}

\newcommand{\snum}{\ensuremath{m}}

\newcommand{\meanvec}{\ensuremath{\nu}}

\newcommand{\Mess}{\ensuremath{M}}

\newcommand{\Bsur}{\ensuremath{B}}
\newcommand{\Bsurd}{\ensuremath{\Bsur^*}}

\newcommand{\Relax}{\ensuremath{\operatorname{REL}}}

\newcommand{\Reg}{\ensuremath{R}}
\newcommand{\Likesur}{\ensuremath{\ell_{\Bsur}}}

\newcommand{\treecol}{\ensuremath{\mathfrak{T}}}

\newcommand{\eparest}{\ensuremath{\estim{\eparam}}}
\newcommand{\meanparest}{\ensuremath{\estim{\meanpar}}}

\newcommand{\convprob}{\stackrel{p}{\longrightarrow}}
\newcommand{\convdist}{\stackrel{d}{\longrightarrow}}

\newcommand{\myLocset}{\ensuremath{\Locset_\clipot}}
\newcommand{\myMargset}{\ensuremath{\Margset_\clipot}}
\newcommand{\myRelaxset}{\ensuremath{\Relax_\clipot}}

\newcommand{\mus}{\ensuremath{\mu^*}}
\newcommand{\MSE}{\operatorname{R}}

\newcommand{\gmean}{\ensuremath{\nu}}
\newcommand{\gvar}{\ensuremath{\sigma}}

\newcommand{\sigmax}{\ensuremath{\sigma_{\operatorname{max}}}}
\newcommand{\Mseinc}{\ensuremath{\Delta \MSE}}

\newcommand{\coupstr}{\ensuremath{\gamma}}

\newcommand{\gaussest}{\ensuremath{g}}
\newcommand{\gaussestone}{\ensuremath{\gaussest_1}}
\newcommand{\gaussestzero}{\ensuremath{\gaussest_0}}

\newcommand{\datavec}{\ensuremath{\gamma}}

\newcommand{\zopt}{\ensuremath{\estim{z}^{\operatorname{opt}}}}
\newcommand{\zapp}{\ensuremath{\estim{z}^{\operatorname{app}}}}

\newcommand{\MSEOPT}{\ensuremath{\MSE^{\operatorname{opt}}}}
\newcommand{\MSEAPP}{\ensuremath{\MSE^{\operatorname{app}}}}

\begin{document}

\ifthenelse{\equal{\doctype}{MP}}
{
\typeout{NO FORMATTING AVAILABLE FOR MP OPTION}

}
%%%%%%%
{
\typeout{(**********************************}
\begin{center}

{{\LARGE \bf{Inconsistent parameter estimation in Markov random
fields: Benefits in the computation-limited setting}}} \\

\vspace*{.5in}

{\large {
\begin{tabular}{c}
Martin J. Wainwright \\
Department of Statistics, and \\
Department of Electrical Engineering and Computer Sciences \\
University of California, Berkeley \\
\texttt{wainwrig@\{eecs,stat\}.berkeley.edu}
\end{tabular}
}}

\vtiny

{\large {February 17, 2006}}

\vtiny

Department of Statistics, UC Berkeley \\
Technical Report 690
\end{center}
}
% end ifthen (otherwise for stattech)

\begin{abstract}

Consider the problem of joint parameter estimation and prediction in a
Markov random field: i.e., the model parameters are estimated on the
basis of an initial set of data, and then the fitted model is used to
perform prediction (e.g., smoothing, denoising, interpolation) on a
new noisy observation.  Working under the restriction of limited
computation, we analyze a joint method in which the \emph{same convex
variational relaxation} is used to construct an M-estimator for
fitting parameters, and to perform approximate marginalization for the
prediction step.  The key result of this paper is that in the
computation-limited setting, using an inconsistent parameter estimator
(i.e., an estimator that returns the ``wrong'' model even in the
infinite data limit) can be provably beneficial, since the resulting
errors can partially compensate for errors made by using an
approximate prediction technique.  En route to this result, we analyze
the asymptotic properties of M-estimators based on convex variational
relaxations, and establish a Lipschitz stability property that holds
for a broad class of variational methods.  We show that joint
estimation/prediction based on the reweighted sum-product algorithm
substantially outperforms a commonly used heuristic based on ordinary
sum-product.
%\footnote{Work partially supported by Intel Corporation Equipment
% Grant 22978, an Alfred P. Sloan Foundation Fellowship, and NSF Grant
% DMS-0528488.}

\end{abstract}

{\fn{
\noindent {\bf Keywords:} Graphical model; Markov random field; belief
propagation; sum-product algorithm; variational method; parameter
estimation; pseudolikelihood; prediction error; Lipschitz stability;
mixture of Gaussian.  }}

\section{Introduction}

Graphical models such as Markov random fields (MRFs) are widely used
in many application domains, including spatial statistics, statistical
signal processing, and communication theory. A fundamental limitation
to their practical use is the infeasibility of computing various
statistical quantities (e.g., marginals, data likelihoods etc.); such
quantities are of interest both Bayesian and frequentist settings.
Sampling-based methods, especially those of the Markov chain Monte
Carlo (MCMC) variety~\cite{Liu,Robert}, represent one approach to
obtaining stochastic approximations to marginals and likelihoods.  A
disadvantage of sampling methods is their relatively high
computational cost.  For instance, in applications with severe limits
on delay and computational overhead (e.g., error-control coding,
real-time tracking, video compression), MCMC methods are likely to be
overly slow.  It is thus of considerable interest for various
application domains to consider less computationally intensive methods
for generating approximations to marginals, log likelihoods, and other
relevant statistical quantities.

Variational methods are one class of techniques that can be used to
generate deterministic approximations in Markov random fields (MRFs).
At the foundation of these methods is the fact that for a broad class
of MRFs, the computation of the log likelihood and marginal
probabilities can be reformulated as a convex optimization problem
(see~\cite{WaiJor05chapter,WaiJor03Monster} for an overview).
Although this optimization problem is intractable to solve exactly for
general MRFs, it suggests a principled route to obtaining
approximations---namely, by relaxing the original optimization
problem, and taking the optimal solutions to the relaxed problem as
approximations to the exact values.  In many cases, optimization of
the relaxed problem can be carried out by ``message-passing''
algorithms, in which neighboring nodes in the Markov random field
convey statistical information (e.g., likelihoods) by passing
functions or vectors (referred to as messages).

Estimating the parameters of a Markov random field from data poses
another significant challenge.  A direct approach---for instance, via
(regularized) maximum likelihood estimation---entails evaluating the
cumulant generating (or log partition) function, which is
computationally intractable for general Markov random fields.  One
viable option is the pseudolikelihood method~\cite{Besag75,Besag77},
which can be shown to produce consistent parameter estimates under
suitable assumptions, though with an associated loss of statistical
efficiency.  Other researchers have studied algorithms for ML
estimation based on stochastic
approximation~\cite{Younes88,Benveniste90}, which again are consistent
under appropriate assumptions, but can be slow to converge.
\subsection{Overview}

\begin{figure}[h]
\bec \psfrag{#ydat#}{{\footnotesize$\{X^i\}$}}
\psfrag{#compatapp#}{$\eparest$} \psfrag{#compatml#}{$\eparams$}
\psfrag{#xhatml#}{{\fn{$\zopt(Y; \eparams)$}}}
\psfrag{#xhatapp#}{{\fn{$\zapp(Y; \eparest)$}}}
\psfrag{#xnew#}{{\fn{$Y$}}} \psfrag{#err#}{{\fn{$\operatorname{Error:}
\; \; \|\zopt(Y; \eparams) - \zapp(Y; \eparest)\|$}}}

\widgraph{0.95\textwidth}{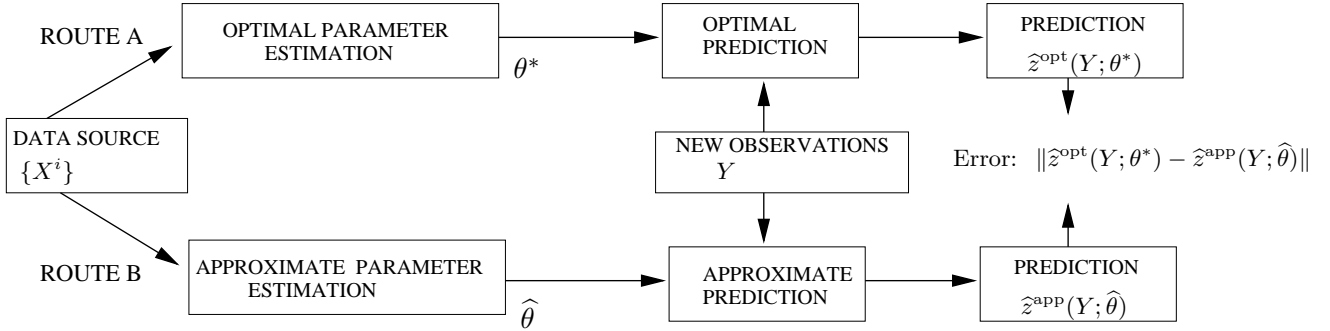}
\enc
\caption{Route A: computationally intractable combination of parameter
estimation and prediction.  Route B: computationally efficient
combination of approximate parameter estimation and prediction.}
\label{FigEstPred}
\end{figure}

As illustrated in Figure~\ref{FigEstPred}, the problem domain of
interest in this paper is that of joint estimation and prediction in a
Markov random field.  More precisely, given samples $\{X^1, \ldots,
X^\dnum \}$ from some unknown underlying model $p(\, \cdot \, ;
\eparams)$, the first step is to form an estimate of the model
parameters.  Now suppose that we are given a noisy observation of a
new sample path $Z \sim p(\, \cdot \, ; \eparams)$, and that we wish
to form a (near)-optimal estimate of $Z$ using the fitted model, and
the noisy observation (denoted $Y$). Examples of such prediction
problems include signal denoising, image interpolation, and decoding
of error-control codes.  Disregarding any issues of computational cost
and speed, one could proceed via Route A in
Figure~\ref{FigEstPred}---that is, one could envisage first using a
standard technique (e.g., regularized maximum likelihood) for
parameter estimation, and then carrying out the prediction step (which
might, for instance, involve computing certain marginal probabilities)
by Monte Carlo methods.

This paper, in contrast, is concerned with the
\emph{computation-limited} setting, in which both sampling or brute
force methods are overly intensive.  With this motivation, a number of
researchers have studied the use of approximate message-passing
techniques, both for problems of
prediction~\cite{Heskes03,Ihler05,Minka01,Tatikonda03,Wainwright01a,Wiegerinck05,Yedidia05}
as well as for parameter
estimation~\cite{Leisink00,Sutton05,Teh03,Wainwright03a}.  However,
despite their wide-spread use, the theoretical understanding of such
message-passing techniques remains limited\footnote{The behavior of
sum-product is relatively well understood in certain settings,
including graphs with single cycles~\cite{Weiss00} and Gaussian
models~\cite{FreemanMAP01,Rus99}.  Similarly, there has been
substantial progress for graphs with high girth~\cite{Richardson01a},
but much of this analysis breaks down in application to graphs with
short cycles}, especially for parameter estimation.  Consequently, it
is of considerable interest to characterize and quantify the loss in
performance incurred by using computationally tractable methods versus
exact methods (i.e., Route B versus A in Figure~\ref{FigEstPred}).
More specifically, our analysis applies to variational methods that
are based \emph{convex relaxations}.  This class includes a broad
range of extant methods---among them the tree-reweighted sum-product
algorithm~\cite{Wainwright05}, reweighted forms of generalized belief
propagation~\cite{Wiegerinck05}, and semidefinite
relaxations~\cite{WaiJor05chapter}. Moreover, it is straightforward to
modify other message-passing methods (e.g., expectation
propagation~\cite{Minka01}) so as to ``convexify'' them.

At a high level, the key idea of this paper is the following: given
that approximate methods can lead to errors at both the estimation and
prediction phases, it is natural to speculate that these sources of
error might be arranged to partially cancel one another.  The
theoretical analysis of this paper confirms this intuition: we show
that with respect to end-to-end performance, it is in fact beneficial,
even in the infinite data limit, to learn the ``wrong'' the model by
using \emph{inconsistent} methods for parameter estimation.  En route
to this result, we analyze the asymptotic properties of M-estimators
based on convex variational relaxations, and establish a Lipschitz
stability property that holds for a broad class of variational
methods.  We show that joint estimation/prediction based on the
reweighted sum-product algorithm substantially outperforms a commonly
used heuristic based on ordinary sum-product.

The remainder of this paper is organized as follows.
Section~\ref{SecBackground} provides background on Markov random
fields and associated variational representations, as well as the
problem statement.  In Section~\ref{SecConvSurr}, we introduce the
notion of a convex surrogate to the cumulant generating function, and
then illustrate this notion via the tree-reweighted Bethe
approximation~\cite{Wainwright05}. In Section~\ref{SecJoint}, we
describe how any convex surrogate defines a particular joint scheme
for parameter estimation and prediction.  Section~\ref{SecAnal}
provides results on the asymptotic behavior of the estimation step, as
well as the stability of the prediction step.
Section~\ref{SecMixGauss} is devoted to the derivation of performance
bounds for joint estimation and prediction methods, with particular
emphasis on the mixture-of-Gaussians observation model.  In
Section~\ref{SecExperiments}, we provide experimental results on the
performance of a joint estimation/prediction method based on the
tree-reweighted Bethe surrogate, and compare it to a heuristic method
based on the ordinary belief propagation algorithm.  We conclude in
Section~\ref{SecDiscussion} with a summary and discussion of
directions for future work.

\section{Background and problem statement}
\label{SecBackground}

\subsection{Multinomial Markov random fields}

Consider an undirected graph $\graph = (\vertex, \edge)$, consisting
of a set of vertices $\vertex = \{1, \ldots, \vnum \}$ and an edge set
$\edge$.  We associate to each vertex $s \in \vertex$ a multinomial
random variable $X_s$ taking values in the set $\statesp_s = \{0,1,
\ldots, \snum-1\}$.  We use the lower case letter $x_s$ to denote
particular realizations of the random variable $X_s$ in the set
$\statesp_s$.  This paper makes use of the following exponential
representation of a pairwise Markov random field over the multinomial
random vector $X \defn \{ X_s, \; s \in \vertex \}$.  We begin by
defining, for each $j = 1, \ldots, \snum -1$, the $\{0,1\}$-valued
indicator function
\begin{eqnarray}
\label{EqnDefnSingInd}
\Ind_j[x_s] & \defn & \begin{cases} 1 & \mbox{if $x_s =j$} \\
0 & \mbox{otherwise} \end{cases}
\end{eqnarray}
These indicator functions can be used to define a potential function
$\eparam_s(\cdot): \statesp_s \rightarrow \real$ via
\begin{eqnarray}
\label{EqnDefnJointInd}
\eparam_s(x_s) & \defn & \sum_{j=1}^{\snum-1} \eparam_{s;j}
\Ind_j[x_s]
\end{eqnarray}
where $\eparam_s = \{\eparam_{s;j}, j = 1, \ldots, \snum - 1\}$ is the
vector of exponential parameters associated with the potential.  Our
exclusion of the index $j=0$ is deliberate, so as to ensure that the
collection of indicator functions $\clipot_s(x_s) \defn \{
\Ind_j[x_s], \; j = 1, \ldots, \snum - 1\}$ remain affinely
independent.  In a similar fashion, we define for any pair $(s,t) \in
\edge$ the pairwise potential function
\begin{eqnarray}
\eparam_{st}(x_s, x_t) & \defn & \sum_{j=1}^{\snum-1} \sum_{k =
1}^{\snum-1} \eparam_{st;jk} \Ind_j[x_s] \; \Ind_k[x_t],
\end{eqnarray}
where we use $\eparam_{st} \defn \{ \eparam_{st;jk}, \; j, k = 1, 2,
\ldots, \snum -1 \}$ to denote the associated collection of
exponential parameters, and $\clipot_{st}(x_s, x_t) \defn \{
\Ind_j[x_s] \, \Ind_k[x_s], \; j,k=1,2, \ldots, \snum - 1 \}$ for the
associated set of sufficient statistics.

Overall, the probability mass function of the multinomial Markov
random field in exponential form can be written as
\begin{eqnarray}
\label{EqnExpFam}
\pd{\eparam}{x} & = & \exp \big \{ \sum_{s \in \vertex} \eparam_s(x_s)
+ \sum_{(s,t) \in \edge} \eparam_{st}(x_s, x_t) - \Partc(\eparam) \big
\}.
\end{eqnarray}
Here the function
\begin{eqnarray}
\Partc(\eparam) & \defn & \log \Big[\sum_{x \in \statesp^\vnum} \exp
\big \{\sum_{s \in \vertex} \eparam_s(x_s) + \sum_{(s,t) \in \edge}
\eparam_{st}(x_s, x_t) \big \} \Big]
\end{eqnarray}
is the logarithm of the normalizing constant associated with
$\pd{\eparam}{\cdot}$.

The collection of distributions thus defined can be viewed as a
regular and minimal exponential family~\cite{Brown86}.  In particular,
the exponential parameter $\eparam$ and the vector of sufficient
statistics $\clipot$ are formed by concatenating the exponential
parameters (respectively indicator functions) associated with each
vertex and edge---viz.
\begin{subequations}
\label{EqnDefnOver}
\begin{eqnarray}
\eparam & = & \{ \eparam_s, s \in \vertex \} \cup \{ \eparam_{st}, \;
(s,t) \in \edge \} \\
\clipot(x) & = & \{ \clipot_s(x_s), s \in \vertex \} \cup \{
\clipot_{st}(x_s, x_t), \; (s,t) \in \edge \} 
\end{eqnarray}
\end{subequations}
This notation allows us to write equation~\eqref{EqnExpFam} more
compactly as $\pd{\eparam}{x}= \exp \{ \inprod{\eparam}{\clipot(x)} -
\Partc(\eparam) \}$.  A quick calculation shows that $\eparam \in
\real^\df$, where $\df = \vnum (\snum-1) + |\edge| \, (\snum-1)^2$ is
the dimension of this exponential family.

\noindent The following properties of $\Partc$ are well-known:
\blems
\label{LemBasic}
\noindent (a) The function $\Partc$ is convex, and strictly so when
the sufficient statistics are affinely independent. \\
\noindent (b) It is an infinitely differentiable function, with
derivatives corresponding to cumulants.  In particular, for any
indices $\sumind, \beta \in \{1, \ldots, \df\}$, we have
\begin{equation}
\pdif{\Partc}{\eparam_\sumind} = \Exs_\eparam[\clipot_\sumind(X)],
\qquad \pdifftwo{\Partc}{\eparam_\sumind}{\eparam_\beta} =
\covv{\clipot_\sumind(X)}{\clipot_\beta(X)}{\eparam},
\end{equation}
where $\Exs_\eparam$ and $\cov_\eparam$ denote the expectation and
covariance respectively.  
\elems
\noindent We use $\meanpar \in \real^\df$ to denote the vector of
\emph{mean parameters} defined element-wise by $\meanpar_\sumind =
\Exs_\eparam[\clipot_\sumind(X)]$ for any $\sumind \in \{1, \ldots,
\df \}$.  A convenient property of the sufficient statistics $\clipot$
defined in equations~\eqref{EqnDefnSingInd}
and~\eqref{EqnDefnJointInd} is that these mean parameters correspond
to marginal probabilities. For instance, when $\alpha = (s;j)$ or
$\alpha = (st;jk)$, we have respectively
\begin{subequations}
\label{EqnOver}
\begin{eqnarray}
\label{EqnOvera}
\meanpar_{s;j} & = & \Exs_\eparam[\Ind_j[x_s]] = \pd{\eparam}{X_s =
j}, \quad \mbox{and} \\
\label{EqnOverb}
\meanpar_{st;jk} & = & \Exs_\eparam \big \{ \Ind_j[x_s]\, \Ind_k[x_t]
\big \} = \pd{\eparam}{X_s = j, X_t =k}.
\end{eqnarray}
\end{subequations}

\section{Construction of convex surrogates}
\label{SecConvSurr}
This section is devoted to a systematic procedure for constructing
convex functions that represent approximations to the cumulant
generating function.  We begin with a quick development of an exact
variational principle, one which is intractable to solve in general
cases.  (More details on this exact variational principle can be found
in the papers~\cite{WaiJor03Monster,WaiJor05chapter}.) Nonetheless, this
exact variational principle is useful, in that various natural
relaxations of the optimization problem can be used to define convex
surrogates to the cumulant generating function.  After a high-level
description of such constructions in general, we then illustrate it
more concretely with the particular case of the ``convexified'' Bethe
entropy~\cite{Wainwright05}.

\subsection{Exact variational representation}

Since $\Partc$ is a convex and continuous function (see
Lemma~\ref{LemBasic}), the theory of convex duality~\cite{Rockafellar}
guarantees that it has a variational representation, given in terms of
its conjugate dual function $\Dualc:\real^\df \rightarrow \real \cup
\{+\infty\}$, of the following form
\begin{eqnarray}
\Partc(\eparam) & = & \sup_{\meanpar \in \real^\df} \big \{
\eparam^T \meanpar - \Dualc(\meanpar) \big \}.
\end{eqnarray}
In order to make effective use of this variational representation, it
remains determine the form of the dual function.  A useful fact is
that the exponential family~\eqref{EqnExpFam} arises naturally as the
solution of an entropy maximization problem.  In particular, consider
the set of linear constraints
\begin{eqnarray}
\label{EqnCon}
\Exs_p[\clipot(X)] \, \defn \; \sum_{x \in \statespn} p(x)
\clipot_\sumind(x) & = & \meanpar_\sumind \quad \mbox{for $\sumind =1,
\ldots, \df$},
\end{eqnarray}
where $\meanpar \in \real^\df$ is a set of target mean parameters.
Letting $\mathcal{P}$ denote the set of all probability distributions
with support on $\statespn$, consider the \emph{constrained entropy
maximization problem}: maximize the entropy $H(p) \defn - \sum_{x \in
\statespn} p(x) \log p(x)$ subject to the constraints~\eqref{EqnCon}.

A first question is when there any distributions $p$ that satisfy the
constraints~\eqref{EqnCon}.  Accordingly, we define the set
\begin{eqnarray}
\myMargset(\graph) & \defn & \big \{ \meanpar \in \real^{\df} \, \big | \,
\meanpar = \Exs_p[\clipot(X)] \quad \mbox{for some $p \in
\mathcal{P}$} \big \},
\end{eqnarray}
corresponding to the set of $\meanpar$ for which the constraint
set~\eqref{EqnCon} is non-empty.  For any \mbox{$\meanpar \notin
\myMargset(\graph)$,} the optimal value of the constrained
maximization problem is $-\infty$ (by definition, since the problem is
infeasible).  Otherwise, it can be shown that the optimum is attained
at a unique distribution in the exponential family, which we denote by
$\pd{\eparam(\meanpar)}{\cdot}$.  Overall, these facts allow us to
specify the conjugate dual function as follows:
\begin{eqnarray}
\label{EqnDefnDualFun}
\Dualc(\meanpar) & = & \begin{cases} -H(\pd{\eparam(\meanpar)}{\cdot})
& \mbox{if $\meanpar \in \myMargset(\graph)$} \\
+\infty & \mbox{otherwise.}
		       \end{cases}
\end{eqnarray}
See the technical report~\cite{WaiJor03Monster} for more details of
this dual calculation.  With this form of the dual function, we are
guaranteed that the cumulant generating function $\Partc$ has the
following variational representation:
\begin{eqnarray}
\label{EqnVarRep}
\Partc(\eparam) & = & \max_{\meanpar \in \myMargset(\graph) } \big \{
\eparam^T \meanpar - \Dualc(\meanpar) \big \}.
\end{eqnarray}
However, in general, solving the variational problem~\eqref{EqnVarRep}
is intractable.  This intractability should not be a surprise, since
the cumulant generating function is intractable to compute for a
general graphical model.  The difficulty arises from two sources.
First, the \emph{constraint set} $\myMargset(\graph)$ is extremely
difficult to characterize exactly for a general graph with cycles.
For the case of a multinomial Markov random field~\eqref{EqnExpFam},
it can be seen (using the Minkowski-Weyl theorem) that
$\myMargset(\graph)$ is a polytope, meaning that it can be
characterized by a finite number of linear constraints.  The question,
of course, is how rapidly this number of constraints grows with the
number of nodes $\vnum$ in the graph.  Unless certain fundamental
conjectures in computational complexity turn out to be false, this
growth must be non-polynomial (see Deza and Laurent~\cite{Deza97} for
an in-depth discussion of the binary case).  Tree-structured graphs
are a notable exception, for which the junction tree
theory~\cite{Lauritzen} guarantees that the growth is only linear in
$\vnum$.

Second, the \emph{dual function} $\Dualc$ lacks a closed-form
representation for a general graph.  Note in particular that the
representation~\eqref{EqnDefnDualFun} is not explicit, since it
requires solving a constrained entropy maximization problem in order
to compute the value $\Ent(\pd{\eparam(\meanpar)}{\cdot})$.  Again,
important exceptions to this rule are tree-structured graphs.  Here a
special case of the junction tree theory guarantees that any Markov
random field on a tree $\treegr = (\vertex, \edge(\treegr))$ can be
factorized in terms of its marginals as follows
\begin{eqnarray}
\label{EqnTreeFact}
\pd{\eparam(\meanpar)}{x} & = & \prod_{s \in \vertex} \meanpar_s(x_s)
\prod_{(s,t) \in \edge(\treegr)} \frac{\meanpar_{st}(x_s,
x_t)}{\meanpar_s(x_s) \meanpar_t(x_t)}.
\end{eqnarray}
Consequently, in this case, the negative entropy (and hence the dual
function) can be computed explicitly as
\begin{eqnarray}
\label{EqnTreeEnt}
-\Dualc(\meanpar; \treegr) & = & \sum_{s \in \vertex}
 \Ent_s(\meanpar_s) - \sum_{(s,t) \in \edge(\treegr)},
 \Info_{st}(\meanpar_{st})
\end{eqnarray}
where $\Ent_s(\meanpar_s) \; \defn \; -\sum_{x_s} \meanpar_s(x_s) \log
\meanpar_s(x_s)$ and $\Info_{st}(\meanpar_{st}) \defn \sum_{x_s, x_t}
\meanpar_{st}(x_s, x_t) \log \frac{\meanpar_{st}(x_s,
x_t)}{\meanpar_s(x_s) \meanpar_t(x_t)}$ are the singleton entropy and
mutual information, respectively, associated with the node $s \in
\vertex$ and edge $(s,t) \in \edge(\treegr)$.  For a general graph
with cycles, in contrast, the dual function lacks such an explicit
form, and is not easy to compute.

Given these challenges, it is natural to consider approximations to
$\Dualc$ and $\myMargset(\graph)$.  As we discuss in the following
section, the resulting relaxed optimization problem defines a convex
surrogate to the cumulant generating function.

\subsection{Convex surrogates to the cumulant generating function}

We now describe a general procedure for constructing convex surrogates
to the cumulant generating function, consisting of two main
ingredients.  Given the intractability of characterizing the marginal
polytope $\myMargset(\graph)$, it is natural to consider a relaxation.
More specifically, let $\myRelaxset(\graph)$ be a convex and compact
set that acts as an outer bound to $\myMargset(\graph)$.  We use
$\taupar$ to denote elements of $\myRelaxset(\graph)$, and refer to
them as \emph{pseudomarginals} since they represent relaxed versions
of local marginals.  The second ingredient is to designed to sidestep
the intractability of the dual function: in particular, let $\Bsurd$
be a strictly convex and twice continuously differentiable
approximation to $\Dualc$.  We require that the domain of $\Bsurd$
(i.e., $\dom(\Bsurd) \defn \{\taupar \in \real^\df \, | \,
\Bsurd(\taupar) < +\infty \}$) be contained within the relaxed
constraint set $\myRelaxset(\graph)$.

By combining these two approximations, we obtain a convex surrogate
$\Bsur$ to the cumulant generating function, specified via the
solution of the following relaxed optimization problem
\begin{eqnarray}
\label{EqnDefnConvSur}
\Bsur(\eparam) & \defn & \max_{\taupar \in \myRelaxset(\graph)} \big
\{\eparam^T \taupar - \Bsurd(\taupar) \big \}.
\end{eqnarray}
Note the parallel between this definition~\eqref{EqnDefnConvSur} and
the variational representation of $\Partc$ in
equation~\eqref{EqnVarRep}.

The function $\Bsur$ so defined has several desirable properties, as
summarized in the following proposition:
\bprops
\label{PropBasicProp}
Any convex surrogate $\Bsur$ defined via~\eqref{EqnDefnConvSur} has
the following properties:
\begin{enumerate}
\item[(i)] For each $\eparam \in \real^\df$, the optimum defining
$\Bsur$ is attained at a unique point $\taupar(\eparam)$.
\item[(ii)] The function $\Bsur$ is convex on $\real^\df$.
\item[(iii)] It is differentiable on $\real^\df$, and more
specifically:
\begin{eqnarray}
\nabla \Bsur(\eparam) & = & \taupar(\eparam).
\end{eqnarray}
\end{enumerate}

\eprops
\spro
\noindent (i) By construction, the constraint set $\myRelaxset(\graph)$ is
compact and convex, and the function $\Bsurd$ is strictly convex, so
that the optimum is attained at a unique point $\taupar(\eparam)$. \\
\noindent (ii) Observe that $\Bsur$ is defined by the maximum of a
collection of functions linear in $\eparam$, which ensures that it is
convex~\cite{Bertsekas_nonlin}. \\
\noindent (iii) Finally, the function $\eparam^T \taupar -
\Bsurd(\taupar)$ satisfies the hypotheses of Danskin's
theorem~\cite{Bertsekas_nonlin}, from which we conclude that $\Bsur$
is differentiable with $\nabla \Bsur (\eparam) = \taupar(\eparam)$ as
claimed.
\fpro

Given the interpretation of $\taupar(\eparam)$ as a pseudomarginal,
this last property of $\Bsur$ is analogous to the well-known cumulant
generating property of $\Partc$---namely, $\nabla \Partc(\eparam) =
\meanpar(\eparam)$---as specified in Lemma~\ref{LemBasic}.

\subsection{Convexified Bethe surrogate}

The following example provides a more concrete illustration of this
constructive procedure, using a tree-based approximation to the
marginal polytope, and a convexifed Bethe entropy
approximation~\cite{Wainwright05}.  As with the ordinary Bethe
approximation~\cite{Yedidia05}, the cost function and constraint set
underlying this approximation are exact for any tree-structured Markov
random field.

\paragraph{Relaxed polytope:}  We begin by describing a relaxed version 
$\myRelaxset(\graph)$ of the marginal polytope
$\myMargset(\clipotvec)$.  Let $\taupar_s$ and $\taupar_{st}$
represent a collection of singleton and pairwise pseudomarginals,
respectively, associated with vertices and edges of a graph $\graph$.
These quantities, as locally valid marginal distributions, must
satisfy the following set of local consistency conditions:
\begin{eqnarray}
\myLocset(\graph) & \defn & \big \{ \taupar \in \real^\df_+ \, \big |
\, \sum_{x_s} \taupar_s(x_s) = 1, \; \; \sum_{x_t} \taupar_{st}(x_s,
x_t) = \taupar_s(x_s) \big \}.
\end{eqnarray}
By construction, we are guaranteed the inclusion $\myMargset(\graph)
\subset \myLocset(\graph)$.  Moreover, a special case of the junction
tree theory~\cite{Lauritzen} guarantees that equality holds when the
underlying graph is a tree (in particular, any $\taupar \in
\myLocset(\graph)$ can be realized as the marginals of the
tree-structured distribution of the form~\eqref{EqnTreeFact}).
However, the inclusion is strict for any graph with cycles; see
Appendix~\ref{AppStrict} for further discussion of this issue.

\paragraph{Entropy approximation:} 
We now define an entropy approximation $\Cbethe$ that is finite for
any pseudomarginal $\taupar$ in the relaxed set $\myLocset(\graph)$.
We begin by considering a collection $\{ \treegr \in \treecol \}$ of
spanning trees associated with the original graph.  Given $\taupar \in
\myLocset(\graph)$, there is---for each spanning tree $\treegr$---a
unique tree-structured distribution that has marginals $\taupar_s$ and
$\taupar_{st}$ on the vertex set $\vertex$ and edge set
$\edge(\treegr)$ of the tree.  Using equations~\eqref{EqnTreeFact}
and~\eqref{EqnTreeEnt}, the entropy of this tree-structured
distribution can be computed explicitly.  The \emph{convexified Bethe
entropy} approximation is based on taking a convex combination of
these tree entropies, where each tree is weighted by a probability
$\rho(\treegr) \in [0,1]$.  Doing so and expanding the sum yields
\begin{eqnarray}
\label{EqnDefnConvBethe}
\Cbethe(\taupar) & \defn & \sum_{\treegr \in \treecol} \rho(\treegr)
\Big \{ \sum_{s \in \vertex} \Ent_s(\taupar_s) - \sum_{(s,t) \in
\edge(\treegr)} \Info_{st}(\taupar_{st}) \Big \} \; = \; \sum_{s \in
\vertex} \Ent_s(\taupar_s) - \sum_{(s,t) \in \edge} \rho_{st}
\Info_{st}(\taupar_{st}),
\end{eqnarray}
where $\rho_{st} = \sum_{\treegr} \rho(\treegr) \Ind[(s,t) \in
\treegr]$ are the \emph{edge appearance probabilities} defined by the
distribution $\rho$ over the tree collection.  By construction, the
function $\Cbethe$ is differentiable; moreover, it can be
shown~\cite{Wainwright05} that it is strictly convex for any vector
$\{\rho_{st} \}$ of strictly positive edge appearance probabilities.

\paragraph{Bethe surrogate and reweighted sum-product:} 
We use these two ingredients---the relaxation $\myLocset(\graph)$ of
the marginal polytope, and the convexified Bethe entropy
approximation~\eqref{EqnDefnConvBethe}---to define the following convex
surrogate
\begin{eqnarray}
\label{EqnBetheSur}
\Cbsur(\eparam) & \defn & \max_{\taupar \in \myLocset(\graph)} \big \{
\eparam^T \taupar - \Cbethe(\taupar) \big \}.
\end{eqnarray}
Since the conditions of Proposition~\ref{PropBasicProp} are satisfied,
we are guaranteed that $\Cbsur$ is convex and differentiable on
$\real^\df$, and moreover that $\nabla \Cbsur(\eparam) =
\taupar(\eparam)$, where (for each $\eparam \in \real^\df$) the
quantity $\taupar(\eparam)$ denotes the unique optimum of
problem~\eqref{EqnBetheSur}.  Perhaps most importantly, the optimizing
pseudomarginals $\taupar(\eparam)$ can be computed efficiently using a
\emph{tree-reweighted variant} of the sum-product message-passing
algorithm~\cite{Wainwright05}. This method operates by passing
``messages'', which in the multinomial case are simply $m$-vectors of
non-negative numbers, along edges of the graph.  We use $\Mess_{ts} =
\{\Mess_{ts}(i), i=0, \ldots, m-1\}$ to represent the message passed
from node $t$ to node $s$.  In the tree-reweighted variant, these
messages are updated according to the following recursion
\begin{eqnarray}
\label{EqnTRWMess}
\Mess_{ts}(x_s) & \leftarrow & \sum_{x_t} \exp \Big \{\eparam_t(x_t)
\frac{\eparam_{st}(x_s, x_t)}{\rho_{st}} \Big \} \frac{\prod_{u \in
\neigh(t) \backslash s} \big[\Mess_{ut}(x_t)\big]^{\rho_{ut}}}
{\big[\Mess_{st}(x_t)\big]^{1-\rho_{st}}}.
\end{eqnarray}
Upon convergence of the updates, the fixed point messages $\Mess^*$
yield the unique global optimum of the optimization
problem~\eqref{EqnBetheSur} via the following equations
\begin{subequations}
\label{EqnTRWMarg}
\begin{eqnarray}
\taupar_s(x_s; \eparam) & \propto & \exp \big\{\eparam_s(x_s) \big \}
\prod_{u \in \neigh(s)} \big[\Mess_{us}(x_s)\big]^{\rho_{us}}, \quad
\mbox{and} \\
\taupar_{st}(x_s, x_t; \eparam) & \propto & \exp \big\{\eparam_s(x_s)
+ \eparam_t(x_t) + \frac{\eparam_{st}(x_s, x_t)}{\rho_{st}} \big \}
\frac{\prod \limits_{u \in \neigh(s)}
\big[\Mess_{us}(x_s)\big]^{\rho_{us}} \;  \prod \limits_{v \in
\neigh(s)} \big[\Mess_{vs}(x_s)\big]^{\rho_{vs}}} { \Mess_{st}(x_t) \;
\Mess_{ts}(x_s) }
\end{eqnarray}
\end{subequations}
Further details on these updates and their properties can be found in
the paper~\cite{Wainwright05}.

\section{Joint estimation and prediction using surrogates}
\label{SecJoint}

We now turn to consideration of how convex surrogates, as constructed
by the procedure described in the previous section, are useful for
both approximate parameter estimation as well as prediction.

\subsection{Approximate parameter estimation}

Suppose that we are given i.i.d. samples $\{X^1, \ldots, X^\dnum \}$
from an MRF of the form~\eqref{EqnExpFam}, where the underlying true
parameter $\eparams$ is unknown.  One standard way in which to
estimate $\eparams$ is via maximum likelihood (possibly with an
additional regularization term); in this particular exponential family
setting, it is straightforward to show that the (normalized) log
likelihood takes the form
\begin{eqnarray}
\label{EqnRegML}
\ell(\eparam) & = & \inprod{\meanparest^\dnum}{\eparam} -
\Partc(\eparam) - \lambda^n \Reg(\eparam)
\end{eqnarray}
where function $\Reg$ is a regularization term with an associated
(possibly data-dependent) weight $\lambda^\dnum$.  The quantities
$\meanparest^\dnum \defn \frac{1}{\dnum} \sum_{i=1}^\dnum
\clipot(X^i)$ are the empirical moments defined by the data.  For the
indicator-based exponential representation~\eqref{EqnOver}, these
empirical moments correspond to a set of singleton and pairwise
marginal distributions, denoted $\meanparest^\dnum_s$ and
$\meanparest^\dnum_{st}$ respectively.

It is intractable to maximize the regularized likelihood directly, due
to the presence of the cumulant generating function $\Partc$.  Thus, a
natural thought is to use the convex surrogate $\Bsur$ to define an
alternative estimator obtained by maximizing the regularized
\emph{surrogate likelihood}:
\begin{eqnarray}
\label{EqnSurLike}
\Likesur(\eparam) & \defn & \inprod{\meanparest^n}{\eparam} -
\Bsur(\eparam) - \lambda^n \Reg(\eparam).
\end{eqnarray}
By design, the surrogate $\Bsur$ and hence the surrogate likelihood
$\Likesur$, as well as their derivatives, can be computed in a
straightforward manner (typically by some sort of message-passing
algorithm).  It is thus straightforward to compute the parameter
$\eparest^\dnum$ achieving the maximum of the regularized surrogate
likelihood (for instance, gradient descent would a simple though naive
method).

For the tree-reweighted Bethe surrogate~\eqref{EqnBetheSur}, we have
shown in previous work~\cite{Wainwright03a} that in the absence of
regularization, the optimal parameter estimates $\eparest^n$ have a
very simple closed-form solution, specified in terms of the weights
$\rho_{st}$ and the empirical marginals $\estim{\meanpar}$.  (We make
use of this closed form in our experimental comparison in
Section~\ref{SecExperiments}; see equation~\eqref{EqnTRWParEst}.)  If
a regularizing term is added, these estimates no longer have a
closed-form solution, but the optimization problem~\eqref{EqnSurLike}
can still be solved efficiently using the tree-reweighted sum-product
algorithm~\cite{Wainwright03a,Wainwright05}.

\subsection{Joint estimation and prediction}
\label{SecJointProcedure}

Using such an estimator, we now consider a joint approach to
estimation and prediction.  Recalling the basic set-up, we are given
an initial set of i.i.d. samples $\{x^1, \ldots, x^\dnum \}$ from
$\pd{\eparams}{\cdot}$, where the true model parameter $\eparams$ is
unknown.  These samples are used to form an estimate of the Markov
random field.  We are then given a noisy observation $y$ of a new
sample $z \sim \pd{\eparams}{\cdot}$, and the goal is to use this
observation in conjunction with the fitted model to form a
near-optimal estimate of $z$.  The key point is that the same convex
surrogate $\Bsur$ is used both in forming the surrogate
likelihood~\eqref{EqnSurLike} for approximate parameter estimation,
and in the variational method~\eqref{EqnDefnConvSur} for performing
prediction.

For a given fitted model parameter $\eparam \in \real^\df$, the
central object in performing prediction is the posterior distribution
$\pd{\eparam}{z \, \mid \, y} \propto \pd{\eparam}{z} \; p(y \, | \,
z)$.  In the exponential family setting, for a fixed noisy observation
$y$, this posterior can always be written as a new exponential family
member, described by parameter $\eparam + \datavec(y)$.  (Here the
term $\datavec(y)$ serves to incorporate the effect of the noisy
observation.)  With this set-up, the procedure consists of the
following steps: \\

\begin{center}
\framebox[0.99 \textwidth]{
\parbox{0.93 \textwidth}{ 
{\bf{Joint estimation and prediction:}}
\begin{enumerate}
\item[1.] Form an approximate parameter estimate $\eparest^\dnum$ from
an initial set of i.i.d. data $\{x^1, \ldots, x^\dnum\}$ by maximizing
the (regularized) surrogate likelihood $\Likesur$.
\item[2.] Given a new noisy observation $y$ (i.e., a contaminated
version of $z \sim \pd{\eparams}{\cdot}$) specified by a factorized
conditional distribution of the form $p(y \, | \, z) =
\prod_{s=1}^\vnum p(y_s \, | \, z_s)$, incorporate it into the model
by forming the new exponential parameter
\begin{equation}
\label{EqnIncData}
\eparest^\dnum_s(\, \cdot \, ) + \datavec_s(y)
\end{equation}
where $\datavec_s(y)$ merges the new data with the fitted model
$\eparest^\dnum$.  (The specific form of $\datavec$ depends on the
observation model.)
\item[3.] Using the message-passing algorithm associated with the
convex surrogate $\Bsur$, compute approximate marginals
$\taupar(\eparest + \datavec)$ for the distribution that combines the
fitted model with the new observation.  Use these approximate
marginals to construct a prediction $\estim{z}(y; \taupar)$
of $z$ based on the observation $y$ and pseudomarginals $\taupar$. \\
%\begin{equation}
%\label{EqnTRWpred}
%\estim{z}_s(y; \taupar) = \sum_{j \in \statesp_s} \taupar_s(j |y;
%\epdata) \Big[ \myvec_j (y_s - \gmean_j) + \gmean_j \Big].
%\end{equation}
\end{enumerate}
}}
\end{center}

\vspace*{.07in} 

Examples of the prediction task in the final step include smoothing
(e.g., denoising of a noisy image) and interpolation (e.g., in the
presence of missing data). We provide a concrete illustration of such
a prediction problem in Section~\ref{SecMixGauss} using a
mixture-of-Gaussians observation model.  The most important property
of this joint scheme is that the \emph{convex surrogate} $\Bsur$
underlies both the parameter estimation phase (used to form the
surrrogate likelihood), and the prediction phase (used in the
variational method for computing approximate marginals).  It is this
matching property that will be shown to be beneficial in terms of
overall performance.

\section{Analysis}
\label{SecAnal}

In this section, we turn to the analysis of the surrogate-based method
for estimation and prediction.  We begin by exploring the asymptotic
behavior of the parameter estimator. We then prove a Lipschitz
stability result applicable to any variational method that is based on
a strongly concave entropy approximation.  This stability result plays
a central role in our subsequent development of bounds on the
performance loss in Section~\ref{SecMixGauss}.

\subsection{Estimator asymptotics}

We begin by considering the asymptotic behavior of the parameter
estimator $\eparest^\dnum$ defined by the surrogate
likelihood~\eqref{EqnSurLike}.  Since this parameter estimator is a
particular type of $M$-estimator, its asymptotic behavior can be
investigated using standard methods, as summarized in the following:
\bprops
\label{PropVan}
Let $\Bsur$ be a strictly convex surrogate to the cumulant generating
function, defined via equation~\eqref{EqnDefnConvSur} with a strictly
concave entropy approximation $-\Bsurd$.  Consider the sequence of
parameter estimates $\{\eparest^\dnum\}$ given by
\begin{eqnarray}
\label{EqnDefnParest}
\eparest^\dnum & \defn & \arg \max_{\eparam \in \real^\df} \left \{
\inprod{\meanparest^n}{\eparam} - \Bsur(\eparam) - \lambda^n
\Reg(\eparam) \right \}
\end{eqnarray}
where $\Reg$ is a non-negative and convex regularizer, and the
regularization parameter satisfies $\lambda^\dnum =
o(\frac{1}{\sqrt{\dnum}})$.

Then for a general graph with cycles, the following results hold:
\begin{enumerate}
\item[(a)] we have $\eparest^n \convprob \eparest$, where $\eparest$
is (in general) distinct from the true parameter $\eparams$.

\item[(b)] the estimator is asymptotically normal:
\begin{eqnarray*}
\sqrt{\dnum} \big[\eparest^\dnum - \eparest \big] & \convdist & N \biggr(0,
\, \big(\nabla^2 \Bsur(\eparest)\big)^{-1} \nabla^2 \Partc(\eparams)
\big(\nabla^2 \Bsur(\eparest)\big)^{-1} \biggr)
\end{eqnarray*}
\end{enumerate}
\eprops
\spro
By construction, the convex surrogate $\Bsur$ and the (negative)
entropy approximation $\Bsurd$ are a Fenchel-Legendre conjugate dual
pair.  From Proposition~\ref{PropBasicProp}, the surrogate $\Bsur$ is
differentiable.  Moreover, the strict convexity of $\Bsur$ and
$\Bsurd$ ensure that the gradient mapping $\nabla \Bsur$ is one-to-one
and onto the relative interior of the constraint set
$\myRelaxset(\graph)$ (see Section 26 of
Rockafellar~\cite{Rockafellar}).  Moreover, the inverse mapping
$(\nabla \Bsur)^{-1}$ exists, and is given by the dual gradient
$\nabla \Bsurd$.

Let $\meanpar^*$ be the moment parameters associated with the true
distribution $\eparams$ (i.e., \mbox{$\meanpar^* =
\Exs_{\eparams}[\clipot(X)]$).}  In the limit of infinite data, the
asymptotic value of the parameter estimate is defined by
\begin{equation}
\label{EqnAsympValue}
\nabla \Bsur(\eparest) = \meanpar^*.
\end{equation}
Note that $\meanpar^*$ belongs to the relative interior of
$\myMargset(\graph)$, and hence to the relative interior of
$\myRelaxset(\graph)$.  Therefore, equation~\eqref{EqnAsympValue} has
a unique solution $\eparest = \nabla^{-1} \Bsur(\meanpar^*)$.

By strict convexity, the regularized surrogate
likelihood~\eqref{EqnDefnParest} has a unique global maximum.  Let us
consider the optimality conditions defining this unique maximum
$\eparest^\dnum$; they are given by $\nabla \Bsur(\eparest^\dnum) =
\meanparest^\dnum - \lambda^n \partial \Reg(\eparest^\dnum)$, where
$\partial \Reg(\eparest^\dnum)$ denotes an arbitrary element of the
subdifferential of the convex function $\Reg$ at the point
$\eparest^\dnum$.  We can now write
\begin{eqnarray}
\label{EqnBasicInq}
\nabla \Bsur(\eparest^\dnum) - \nabla \Bsur(\eparest) & = &
\left[\meanparest^\dnum - \meanpar^* \right] - \lambda^n \partial
\Reg(\eparest^\dnum).
\end{eqnarray}
Taking inner products with the difference $\eparest^\dnum - \eparest$
yields
\begin{equation}
\label{EqnInqOne}
0 \; \stackrel{(a)}{\leq} \; \left[\Bsur(\eparest^\dnum) - \nabla
\Bsur(\eparest)\right]^T \; \left[\eparest^\dnum - \eparest \right] \;
\leq \; \left[\meanparest^\dnum - \meanpar^* \right]^T
\left[\eparest^\dnum - \eparest \right] + \lambda^n \partial
\Reg(\eparest^\dnum)^T \left[\eparest - \eparest^\dnum \right],
\end{equation}
where inequality (a) follows from the convexity of $\Bsur$.  From the
convexity and non-negativity of $\Reg$, we have
\[
\lambda^\dnum \partial \Reg(\eparest^\dnum)^T \left[\eparest -
\eparest^\dnum \right] \; \leq \; \lambda^\dnum \left[ \Reg(\eparest)
- \Reg(\eparest^\dnum)\right] \; \leq \; \lambda^\dnum \Reg(\eparest).
\]
Applying this inequality and Cauchy-Schwartz to
equation~\eqref{EqnInqOne} yields
\begin{equation}
\label{EqnInqTwo}
0 \; \leq \; \left[\Bsur(\eparest^\dnum) - \nabla
\Bsur(\eparest)\right]^T \; \left[\frac{\eparest^\dnum -
\eparest}{\|\eparest^\dnum - \eparest\|} \right] \; \leq \;
\|\meanparest^\dnum - \meanpar^* \| + \lambda^n \Reg(\eparest)
\end{equation}
Since $\lambda^n = o(1)$ by assumption and $\|\meanparest^\dnum -
\meanpar^*\| = o_p(1)$ by the weak law of large numbers, the quantity
$\left[\Bsur(\eparest^\dnum) - \nabla \Bsur(\eparest)\right]^T \;
\left[\frac{\eparest^\dnum - \eparest}{\|\eparest^\dnum - \eparest\|}
\right]$ converges in probability to zero.  By the strict convexity of
$\Bsur$, this fact implies that $\eparest^\dnum$ converges in
probability to $\eparest$, thereby completing the proof of part (a).

To establish part (b), we observe that $\sqrt{\dnum}
\left[\meanparest^\dnum - \meanpar^* \right] \convdist N(0, \nabla^2
\Parta(\eparams))$ by the central limit theorem.  Using this fact and
applying the delta method to equation~\eqref{EqnBasicInq} yields that
\[
\sqrt{\dnum} \nabla^2 \Bsur(\eparest) \; \left[\eparest^\dnum -
\eparest \right] \convdist N \left(0, \nabla^2 \Partc(\eparams)
\right),
\]
where we have used the fact that $\sqrt{\dnum} \lambda^\dnum = o(1)$.
The strict convexity of $\Bsur$ guarantees that $\nabla^2
\Bsur(\eparest)$ is invertible, so that claim (b) follows.
\fpro

A key property of the estimator is its \emph{inconsistency}---i.e.,
the estimated model differs from the true model $\eparams$ even in the
limit of large data. Despite this inconsistency, we will see that the
approximate parameter estimates $\eparest^\dnum$ are nonetheless
useful for performing prediction.

\subsection{Algorithmic stability}

\begin{figure}[h]
\begin{center}
\widgraph{0.35\textwidth}{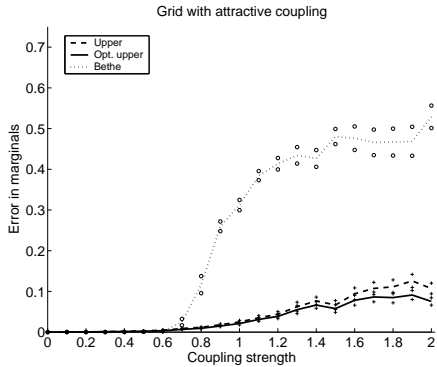}
\end{center}
\caption{Contrast of the instability of the ordinary sum-product
algorithm with the stability of the tree-reweighted
version~\cite{Wainwright05}.  Results shown with a grid with $\vnum =
100$ nodes over a range of attractive coupling strengths.  The
ordinary sum-product undergoes a phase transition, after which the
quality of marginal approximations degrades substantially.  The
tree-reweighted algorithm, shown for two different settings of the
edge weights $\rho_{st}$, remains stable over the full range of
coupling strengths.  See Wainwright et al.~\cite{Wainwright05} for
full details.}
\label{FigUnstable}
\end{figure}

A desirable property of any algorithm---particularly one applied to
statistical data---is that it exhibit an appropriate form of stability
with respect to its inputs.  Not all message-passing algorithms have
such stability properties.  For instance, the standard sum-product
message-passing algorithm, although stable for relatively weakly
coupled MRFs~\cite{Ihler05,Tatikonda02}, can be highly unstable in
other regimes due to the appearance of multiple local optima in the
non-convex Bethe problem.  However, previous experimental work has
shown that methods based on convex relaxations, including the
reweighted sum-product (or belief propagation)
algorithm~\cite{Wainwright03a}, reweighted generalized
BP~\cite{Wiegerinck05}, and log-determinant
relaxations~\cite{WaiJor05chapter} appear to be very stable.  For
instance, Figure~\ref{FigUnstable} provides a simple illustration of
the instability of the ordinary sum-product algorithm, contrasted with
the stability of the tree-reweighted updates.
Wiegerinck~\cite{Wiegerinck05} provides similar results for reweighted
forms of the generalized belief propagation.
Here we provide theoretical support for these empirical observations:
in particular, we prove that, in sharp contrast to non-convex methods,
any variational method based on a strongly convex entropy
approximation is globally stable.  This stability property plays a
fundamental role in providing a performance guarantee on joint
estimation/prediction methods.

We begin by noting that for a multinomial Markov random
field~\eqref{EqnExpFam}, the computation of the exact marginal
probabilities is a globally Lipschitz operation:
\blems
\label{LemExact}
For any discrete Markov random field~\eqref{EqnExpFam}, there is a
constant $L < +\infty$ such that
\begin{eqnarray}
\|\meanpar(\eparam+ \delta) - \meanpar(\eparam) \| & \leq & L \| \delta
\| \qquad \quad \mbox{for all $\eparam, \delta \in \real^\df$.}
\end{eqnarray}
\elems
\noindent This lemma, which is proved in Appendix~\ref{AppExact},
guarantees that small changes in the problem parameters---that is,
``perturbations'' $\delta$---lead to correspondingly small changes in
the computed marginals.

Our goal is to establish analogous Lipschitz properties for
variational methods.  In particular, it turns out that any variational
method based on a suitably concave entropy approximation satisfies
such a stability condition.  More precisely, a function $f:\real^n
\rightarrow \real$ is \emph{strongly convex} if there exists a
constant $c > 0$ such that $f(y) \geq f(x) + \nabla f(x)^T \, \big(y
-x) + \frac{c}{2} \|y - x\|^2$ for all $x, y \in \real^n$.  For a
twice continuously differentiable function, this condition is
equivalent to having the eigenspectrum of the Hessian $\nabla^2 f(x)$
be uniformly bounded below by $c$.  With this definition, we have:
\bprops
\label{PropLipStable}
Consider any strictly convex surrogate $\Bsur$ based on a strongly
concave entropy approximation $-\Bsurd$.  Then there exists a constant
$R < +\infty$ such that
\begin{eqnarray*}
\|\taupar(\eparam+ \delta) - \taupar(\eparam) \| & \leq & R \| \delta
\| \qquad \quad \mbox{for all $\eparam, \delta \in \real^\df$.}
\end{eqnarray*}
\spro
From Proposition~\ref{PropBasicProp}, we have $\tauvec(\eparam) =
\nabla \Bsur(\eparam)$, so that the statement is equivalent to the
assertion that the gradient $\nabla \Bsur$ is a Lipschitz function.
Applying the mean value theorem to $\nabla \Bsur$, we can write
$\nabla \Bsur(\eparam + \delta) - \nabla \Bsur(\eparam) = \nabla^2
\Bsur(\eparam + t \delta) \delta$ where $t \in [0,1]$.  Consequently,
in order to establish the Lipschitz condition, it suffices to show
that the spectral norm of $\nabla^2 \Bsur(\gamma)$ is uniformly
bounded above over all $\gamma \in \real^\df$.  Since $\Bsur$ and
$\Bsurd$ are a strictly convex Legendre pair, we have $\nabla^2
\Bsur(\eparam) = [\nabla^2 \Bsurd(\taupar(\eparam)) ]^{-1}$.  By the
strong convexity of $\Bsurd$, we are guaranteed that the spectral norm
of $\nabla^2 \Bsurd(\taupar)$ is uniformly bounded away from zero,
which yields the claim.
\fpro
\eprops
Many existing entropy approximations can be shown to be strongly
concave.  In Appendix~\ref{AppConv}, we provide a detailed proof of
this fact for the convexified Bethe entropy~\eqref{EqnDefnConvBethe}.
\blems
\label{LemConv}
For any set $\{\rho_{st}\}$ of strictly positive edge appearance
probabilities, the convexified Bethe entropy~\eqref{EqnDefnConvBethe}
is strongly concave.
\elems
\noindent We note that the same argument can be used to establish
strong concavity for the reweighted Kikuchi approximations studied by
Wiegerinck~\cite{Wiegerinck05}.  Moreover, it can be shown that the
Gaussian-based log-determinant relaxation considered in Wainwright and
Jordan~\cite{WaiJor05b} is also strongly concave.  For all of these
variational methods, then, Proposition~\ref{PropLipStable} guarantees
that the pseudomarginal computation is globally Lipschitz stable,
thereby providing theoretical confirmation of previous experimental
results~\cite{Wiegerinck05,Wainwright05,WaiJor05b}.

%%%%%%%%%%%%%%%%%%%%%%%%%%%%%%%%%%%%%%%%%%%%%%%%%%%%%%%%%%%%%%%%

\section{Performance bounds}
\label{SecMixGauss}

In this section, we develop theoretical bounds on the performance loss
of our approximate approach to joint estimation and prediction,
relative to the unattainable Bayes optimum.  So as not to
unnecessarily complicate the result, we focus on the performance loss
in the infinite data limit\footnote{Note, however, that modified forms
of the results given here, modulo the usual $\order(1/\dnum)$
corrections, hold for the finite data setting.} (i.e., for which the
number of samples $\dnum = +\infty$).

In the infinite data setting, the Bayes optimum is unattainable for
two reasons:
\begin{enumerate}
\item it is based on knowledge of the exact parameter $\eparams$, which
is not easy to obtain.
\item it assumes (in the prediction phase) that computing exact
marginal probabilities $\meanpar$ of the Markov random field is
feasible.
\end{enumerate}
Of these two difficulties, it is the latter assumption---regarding the
computation of marginal probabilities---that is the most serious.  As
discussed earlier, there do exist computationally tractable estimators
of $\eparams$ that are consistent though not statistically efficient
under appropriate conditions; one example is the pseudolikelihood
method~\cite{Besag75,Besag77} mentioned previously.  On the other
hand, MCMC methods may be used to generate stochastic approximations
to marginal probabilities, but may require greater than polynomial
complexity.

Recall from Proposition~\ref{PropVan} that the parameter estimator
based on the surrogate likelihood $\Likesur$ is \emph{inconsistent},
in the sense that the parameter vector $\eparest$ returned in the
limit of infinite data is generally not equal to the true parameter
$\eparams$.  Our analysis in this section will demonstrate that this
inconsistency is beneficial.

\subsection{Problem set-up}

Although the ideas and techniques described here are more generally
applicable, we focus here on a special observation model so as to
obtain a concrete result.  

\paragraph{Observation model:}  In particular, we assume that 
the multinomial random vector $X = \{X_s, \; s \in \vertex \}$ defined
by the Markov random field~\eqref{EqnExpFam} is a label vector for the
components in a finite mixture of Gaussians. For each node $s \in
\vertex$, we specify a new random variable $Z_s$ by the conditional
distribution
\[
p(Z_s = z_s \, | \, X_s = j) \sim N(\gmean_j, \gvar^2_j) \quad
\mbox{for $j \in \{0,1, \ldots, m-1\}$},
\]
so that $Z_s$ is a mixture of $m$ Gaussians.  Such Gaussian mixture
models are widely used in spatial statistics as well as statistical
signal and image processing~\cite{Crouse98,Ripley81,Titterington86}.

Now suppose that we observe a noise-corrupted version of $z_s$---
namely, a vector $Y$ of observations with components of the form
\begin{eqnarray}
\label{EqnObsModel}
Y_s = \alsnr Z_s + \sqrt{1- \alsnr^2} W_s,
\end{eqnarray}
where $W_s \sim N(0, 1)$ is additive Gaussian noise, and the parameter
$\alsnr \in [0,1]$ specifies the signal-to-noise ratio (SNR) of the
observation model.  Note that $\alsnr = 0$ corresponds to pure noise,
whereas $\alsnr = 1$ corresponds to completely uncorrupted
observations.  

\paragraph{Optimal prediction:}  Our goal is to compute an optimal 
estimate $\estim{z}(y)$ of $z$ as a function of the observation $Y=y$,
using the mean-squared error as the risk function.  The essential
object in this computation is the posterior distribution
$\pd{\eparams}{x \, \mid \, y} = \pd{\eparams}{x} \, p(y \, \mid \, x)$,
where the conditional distribution $p(y \, \mid  \, x)$ is defined by
the observation model~\eqref{EqnObsModel}.  As shown in the sequel,
the posterior distribution (with $y$ fixed) can be expressed as an
exponential family member of the form $\eparams + \datavec(y)$ (see
equation~\eqref{EqnDefnEpy}).  Disregarding computational cost, it is
straightforward to show that the optimal Bayes least squares estimator
(BLSE) takes the form
\begin{eqnarray}
\label{EqnBayesOpt}
\zopt_s(Y; \eparams) & \defn & \sum_{j=0}^{m-1} \meanpar_s(j; \eparams
+ \datavec(Y)) \biggr[ \myvec_j(\alsnr) \big(Y_s - \gmean_j \big) +
\gmean_j \biggr],
\end{eqnarray}
where $\meanpar_s(j; \eparams + \datavec)$ denotes the marginal
probability associated with the posterior distribution
\mbox{$\pd{\eparams+\datavec}{x}$}, and
\begin{eqnarray}
\label{EqnDefnMyvec}
\myvec_j(\alsnr) & \defn & \frac{\alsnr \sigma^2_j}{\alsnr^2
\sigma^2_j + (1-\alsnr^2)}
\end{eqnarray}
is the usual BLSE weighting for a Gaussian with variance $\gvar^2_j$.

\paragraph{Approximate prediction:} 

Since the marginal distributions $\meanpar_s(j; \eparams + \datavec)$
are intractable to compute exactly, it is natural to consider an
approximate predictor, based on a set $\taupar$ of pseudomarginals
computed from a variational relaxation.  More explicitly, we run the
variational algorithm on the parameter vector $\eparest + \datavec$
that is obtained by combining the new observation $y$ with the fitted
model $\eparest$, and use the outputted pseudomarginals
$\taupar_s(\cdot; \eparest + \datavec)$ as weights in the approximate
predictor
\begin{eqnarray}
\label{EqnAppPred}
\zapp_s(Y; \eparest) & \defn & \sum_{j=0}^{m-1} \taupar_s(j; \eparest
+ \datavec(Y)) \biggr[ \myvec_j(\alsnr) \big(Y_s - \gmean_j \big) +
\gmean_j \biggr],
\end{eqnarray}
where the weights $\myvec$ are defined in
equation~\eqref{EqnDefnMyvec}.

We now turn to a comparison of the Bayes least-squares estimator
(BLSE) defined in equation~\eqref{EqnBayesOpt} to the surrogate-based
predictor~\eqref{EqnAppPred}.  Since (by definition) the BLSE is
optimal for the mean-squared error (MSE), using the surrogate-based
predictor will necessarily lead to a larger MSE.  Our goal is to prove
an upper bound on the maximal possible increase in this MSE, where the
bound is specified in terms of the underlying model $\eparams$ and the
SNR parameter $\alsnr$.  More specifically, for a given problem, we
define the mean-squared errors
\begin{equation}
\MSEAPP(\alsnr, \eparams) \defn \frac{1}{\vnum} \Exs \| \zopt(Y;
\eparams) - Z \|^2, \quad \mbox{and} \quad
\MSEOPT(\alsnr, \eparest) \defn \frac{1}{\vnum} \Exs \| \zapp(Y;
\eparest) - Z \|^2,
\end{equation}
of the Bayes-optimal and surrogate-based predictors, respectively.  We
seek upper bounds on the increase $\Mseinc(\alsnr, \eparams, \eparest)
\defn \MSEAPP(\alsnr, \eparest) - \MSEOPT(\alsnr, \eparams)$ of the
approximate predictor relative to Bayes optimum.

\subsection{Role of stability}

Before providing a technical statement and proof, we begin with some
intuition underlying the bounds, and the role of Lipschitz stability.
First, consider the low SNR regime ($\alsnr \approx 0$) in which the
observation $Y$ is heavily corrupted by noise.  In the limit $\alsnr =
0$, the new observations are pure noise, so that the prediction of $Z$
should be based simply on the estimated model---namely, the true model
$\pd{\eparams}{\cdot}$ in the Bayes optimal case, and the
``incorrect'' model $\pd{\eparest}{\cdot}$ for the method based on
surrogate likelihood.  The key point here is the following: by
properties of the MLE and surrogate-based estimator, the following
equalities hold:
\begin{equation}
\nabla \Partc(\eparams) \; \stackrel{(a)}{=} \; \meanpar(\eparams) \;
\stackrel{(b)}{=} \; \mus \; \stackrel{(c)}{=} \; \taupar(\eparest) \;
\stackrel{(d)}{=} \; \nabla \Bsur(\eparest).
\end{equation}
Here equality (a) follows from Lemma~\ref{LemBasic}, whereas equality
(b) follows from the moment-matching property of the MLE in
exponential families.  Equalities (c) and (d) hold from the
Proposition~\ref{PropBasicProp} and the pseudomoment-matching property
of the surrogate-based parameter estimator (see proof of
Proposition~\ref{PropVan}).  As a key consequence, it follows that the
combination of surrogate-based estimation and prediction is
\emph{functionally indistinguishable} from the Bayes-optimal behavior
in the limit of $\alsnr = 0$.  More specifically, in the limiting
case, the errors systematically introduced by the inconsistent
learning procedure are cancelled out exactly by the approximate
variational method for computing marginal distributions.  Of course,
exactness for $\alsnr = 0$ is of limited interest; however, when
combined with the Lipschitz stability ensured by
Proposition~\ref{PropLipStable}, it allows us to gain good control of
the low SNR regime.  At the other extreme of high SNR ($\alsnr \approx
1$), the observations are nearly perfect, and hence dominate the
behavior of the optimal estimator.  More precisely, for $\alsnr$ close
to $1$, we have $\myvec_j(\alsnr) \approx 1$ for all $j = 0,1, \ldots,
\statenum-1$, so that $\zopt(Y; \eparams) \approx Y \approx \zapp(Y;
\eparest)$.  Consequently, in the high SNR regime, accuracy of the
marginal computation has little effect on the accuracy of the
predictor.

\subsection{Bound on performance loss}

Although bounds of this nature can be developed in more generality,
for simplicity in notation we focus here on the case of $\statenum =
2$ mixture components. We begin by introducing the factors that play a
role in our bound on the performance loss $\Mseinc(\alsnr, \eparams,
\eparest)$. First, the Lipschitz stability enters in the form of the
quantity:
\begin{eqnarray}
\label{EqnDefnLcon}
\Lcon(\eparams; \eparest) & \defn & \sup_{\delta \in \real^\df}
\sigmax\big ( \nabla^2 \Partc(\eparams + \delta) - \nabla^2
\Bsur(\eparest + \delta) \big),
\end{eqnarray}
where $\sigmax$ denotes the maximal singular value.  Following the
argument in the proof of Proposition~\ref{PropLipStable}, it can be
seen that $\Lcon(\eparams; \eparest)$ is finite.

Second, in order to apply the Lipschitz stability result, it is
convenient to express the effect of introducing a new observation
vector $y$, drawn from the additive noise observation
model~\eqref{EqnObsModel}, as a perturbation of the exponential
parameterization.  In particular, for any parameter $\eparam \in
\real^\df$ and observation $y$ from the model~\eqref{EqnObsModel}, the
conditional distribution $p(x \, | \, y; \eparam)$ can be expressed as
$p(x; \eparam + \epy(y, \alsnr))$, where the exponential parameter
$\epy(y, \alsnr)$ has components
\begin{subequations}
\begin{eqnarray}
\label{EqnDefnEpy}
\epy_{s} & = & \frac{1}{2} \Biggr \{ \log \frac{\alsnr^2 \sigma_0^2 +
(1-\alsnr^2)}{\alsnr^2 \sigma^2_1 + (1-\alsnr^2)} \; + \; \frac{(y_s -
\alsnr \meanvec_0)^2}{\alsnr^2 \sigma^2_0 + (1-\alsnr^2)} -\frac{(y_s
- \alsnr \meanvec_1)^2}{\alsnr^2 \sigma^2_1 + (1-\alsnr^2)} \Biggr \}
\qquad \forall \; s \in \vertex. \\
\epy_{st} & = & 0 \qquad \qquad \qquad \qquad \forall \; (s,t) \in \edge.
\end{eqnarray}
\end{subequations}
See Appendix~\ref{AppGamForm} for a derivation of these relations.

Third, it is convenient to have short notation for the Gaussian
estimators of each mixture component:
\begin{eqnarray}
\gaussest_j(Y_s; \alsnr) & \defn & \myvec_j(\alsnr)\, \left(Y_s -
\meanvec_j \right) + \meanvec_j, \qquad \mbox{for $j=0,1$}
\end{eqnarray}

With this notation, we have the following
\btheos
\label{ThmBound}
The MSE increase \mbox{$\Mseinc(\alsnr, \eparams, \eparest) \defn
\MSE(\alsnr, \eparest) - \MSE(\alsnr, \eparams)$} is upper bounded by
\begin{eqnarray}
\label{EqnKeyBound}
\Mseinc(\alsnr, \eparams, \eparest) & \leq & \Exs \left \{ \min \left
(1, \; \Lcon(\eparams; \eparest) \frac{\|\datavec(\Ysca;
\alsnr)\|_2}{\sqrt{\vnum}} \right ) \; \sqrt { \frac{\sum_{s=1}^\vnum
\left |\gaussestone(Y_s) - \gaussestzero(Y_s) \right |^4}{\vnum}}
\right \}.
\end{eqnarray}
\etheos
\noindent Before proving the bound~\eqref{EqnKeyBound}, we begin by
considering its behavior in a few special cases.

\paragraph{Effect of SNR:}  First, consider the low SNR limit in which 
$\alsnr \rightarrow 0^+$.  In this limit, it can be seen that
$\|\epy(\Ysca; \alsnr)\| \rightarrow 0$, so that the the overall bound
$\Mseinc(\alsnr)$ tends to zero.  Similarly, in the high SNR limit as
$\alsnr \rightarrow 1^{-}$, we see that $\myvec_j(\alsnr) \rightarrow
1$ for $j=0,1$, which drives the differences $|\gaussestone(Y_s) -
\gaussestzero(Y_s)|$, and in turn the overall bound $\Mseinc(\alsnr)$
to zero.  Thus, the surrogate-based method is optimal in both the low
and high SNR regimes; its behavior in the intermediate regime is
governed by the balance between these two terms.

\paragraph{Effect of equal variances:}  Now consider the special
case of equal variances $\sigma^2 \equiv \sigma_0^2 = \sigma_1^2$, in
which case $\myvec(\alsnr) \equiv \myvec_0(\alsnr) =
\myvec_1(\alsnr)$.  Thus, the difference $ \gaussestone(Y_s, \alsnr) -
\gaussestzero(Y_s, \alsnr)$ simplifies to $(1 -\myvec(\alsnr)) \,
(\gmean_1 - \gmean_0)$, so that the bound~\eqref{EqnKeyBound} reduces
to
\begin{eqnarray}
\label{EqnAltEqvar}
\Mseinc(\alsnr, \eparams, \eparest) & \leq & (1 -\myvec(\alsnr))^2 \,
(\gmean_1 - \gmean_0)^2 \; \Exs \left \{ \min \left (1, \;
\Lcon(\eparams; \eparest) \frac{\|\datavec(\Ysca;
\alsnr)\|_2}{\sqrt{\vnum}} \right ) \right \}.
\end{eqnarray}
As shown by the simpler expression~\eqref{EqnAltEqvar}, for $\gmean_1
\approx \gmean_0$, the MSE increase is very small, since such a
two-component mixture is close to a pure Gaussian.

\paragraph{Effect of mean difference:}  Finally consider the
case of equal means $\meanvec \equiv \meanvec_0 = \meanvec_1$ in the
two Gaussian mixture components.  In this case, we have
$\gaussestone(Y_s, \alsnr) - \gaussestzero(Y_s, \alsnr) =
[\myvec_1(\alsnr) - \myvec_0(\alsnr)] \; [Y_s - \meanvec ]$, so that
the bound~\eqref{EqnKeyBound} reduces to
\begin{eqnarray}
\label{EqnAltEqmean}
\Mseinc(\alsnr, \eparams, \eparest) & \leq & \left[ \myvec_1(\alsnr) -
\myvec_0(\alsnr) \right]^2 \; \Exs \left \{ \min \left (1, \;
\Lcon(\eparams; \eparest) \frac{\|\epy(\Ysca;
\alsnr)\|_2}{\sqrt{\vnum}} \right ) \sqrt{\frac{\sum_s (\Ysca_s -
\meanvec)^4}{\vnum} } \right \}.
\end{eqnarray}
Here the MSE increase depends on the SNR $\alsnr$ and the difference
\begin{eqnarray*}
\myvec_1(\alsnr) - \myvec_0(\alsnr) & = & \frac{\alsnr
\sigma^2_1}{\alsnr^2 \sigma^2_1 + (1-\alsnr^2) } - \frac{\alsnr
\sigma^2_0}{\alsnr^2 \sigma^2_0 + (1-\alsnr^2)} \; = \;
\frac{(1-\alsnr^2) \; (\sigma_1^2 - \sigma_0^2)}{\left[\alsnr^2
\sigma^2_0 + (1-\alsnr^2)\right] \; \left[\alsnr^2 \sigma^2_1 +
(1-\alsnr^2)\right]}.
\end{eqnarray*}
Observe, in particular, that the MSE increases tends to zero as the
difference $\sigma_1^2 - \sigma_0^2$ decreases.

\subsection{Proof of Theorem~\ref{ThmBound}}

By the Pythagorean relation that characterizes the Bayes least squares
estimator $\zmuvecs{*}{\Ysca}$, we have
\begin{eqnarray*}
\Mseinc(\alsnr; \eparams, \eparest) & \defn & \frac{1}{\vnum} \Exs \|
 \zapp(Y; \eparest) - Z \|_2^2 - \frac{1}{\vnum} \Exs \|\zopt(Y;
 \eparams) - Z \|_2^2 \\
& = & \frac{1}{\vnum} \Exs \| \zapp(Y; \eparest) - \zopt(Y; \eparams)
\|_2^2.
\end{eqnarray*}
Using the definitions of $\zapp(Y; \eparest)$ and $\zopt(Y;
\eparams)$, some algebraic manipulation yields
\begin{eqnarray*}
\left[\zapp_s(Y; \eparest) - \zopt_s(Y; \eparams) \right]^2 & = &
\left[\tau_s(\eparest+\datavec) - \mu_s(\eparams + \datavec) \right]^2
\; \left [ \gaussestone(Y_s) - \gaussestzero(Y_s) \right]^2 \\
& \leq & \left |\tau_s(\eparest + \datavec) - \mu_s(\eparams+
\datavec) \right | \left [ \gaussestone(Y_s) - \gaussestzero(Y_s)
\right]^2,
\end{eqnarray*}
where the second inequality uses the fact that $|\tau_s - \mu_s|
\leq 1$ since $\tau_s$ and $\mu_s$ are marginal probabilities.  Next
we write
\begin{eqnarray}
\frac{1}{\vnum} \| \zapp(Y; \eparest) - \zopt(Y; \eparams) \|_2^2
& \leq & \frac{1}{\vnum} \; \sum_{s=1}^\vnum \left |\tau_s(\eparest +
\datavec) - \mu(\eparams + \datavec) \right | \left [
\gaussestone(Y_s) - \gaussestzero(Y_s) \right]^2 \nonumber \\
\label{EqnInter}
& \leq & \frac{1}{\sqrt{\vnum}} \|\tau(\eparest + \datavec) -
\mu(\eparams + \datavec) \|_2 \; \sqrt { \frac{\sum_{s=1}^\vnum \left
|\gaussestone(Y_s) - \gaussestzero(Y_s) \right |^4}{\vnum}},
\end{eqnarray}
where the last line uses the Cauchy-Schwarz inequality.

It remains to bound the 2-norm $\|\tau(\eparest + \datavec) -
\mu(\eparams + \datavec) \|_2$.  An initial naive bound follows from
the fact $\tau_s, \mu_s \in [0,1]$ implies that $|\tau_s - \mu_s| \leq
1 $, whence
\begin{eqnarray}
\label{EqnMbound1}
\frac{1}{\sqrt{\vnum}} \| \tau - \mu \|_2 \leq & 1.
\end{eqnarray}
An alternative bound, which will be better for small perturbations
$\datavec$, can be obtained as follows.  Using the relation
$\taupar(\eparest) = \mu(\eparams)$ guaranteed by the definition of
the ML estimator and surrogate estimator, we have
\begin{eqnarray*}
\|\tau(\eparest + \datavec) - \mu(\eparams + \datavec)\|_2 & = & \left
\| \left [\tau(\eparest + \datavec) - \tau(\eparest)\right] + \left
[\mu(\eparams) - \mu(\eparams + \datavec)\right] \right \|_2 \\
& = & \left \| \left [ \nabla^2 \Bsur(\eparest + s \datavec) -
\nabla^2 \Partc(\eparams + t \datavec) \right] \datavec \right \|_2,
\end{eqnarray*}
for some $s, t \in [0,1]$, where we have used the mean value theorem.
Thus, using the definition~\eqref{EqnDefnLcon} of $\Lcon$, we have
\begin{eqnarray}
\label{EqnMbound2}
\frac{1}{\sqrt{\vnum}} \; \|\tau(\eparest + \datavec) - \mu(\eparams +
\datavec)\|_2 & \leq & \Lcon(\eparams; \eparest)
\frac{\|\datavec(\Ysca; \alsnr)\|_2}{\sqrt{\vnum}}.
\end{eqnarray}
Combining the bounds~\eqref{EqnMbound1} and~\eqref{EqnMbound2} and
applying them to equation~\eqref{EqnInter}, we obtain
\begin{eqnarray*}
\frac{1}{\vnum} \| \zapp(Y; \eparest) - \zopt(Y; \eparams) \|_2^2 &
\leq & \min \left \{1, \; \Lcon(\eparams; \eparest)
\frac{\|\datavec(\Ysca; \alsnr)\|_2}{\sqrt{\vnum}} \right \} \; \sqrt
{ \frac{\sum_{s=1}^\vnum \left |\gaussestone(Y_s) - \gaussestzero(Y_s)
\right |^4}{\vnum}}.
\end{eqnarray*}
Taking expectations of both sides yields the result. \hfill \qed

%%%%%%%%%%%%% END OF PROOF %%%%%%%%%%%%%%%%%%%%%%%%%%%%%%%%%%%%%%%%%

\section{Experimental results}
\label{SecExperiments}

In order to test our joint estimation/prediction procedure, we have
applied it to coupled Gaussian mixture models on different graphs,
coupling strengths, observation SNRs, and mixture distributions.  Here
we describe both experimental results to quantify the performance loss
of the tree-reweighted sum-product algorithm~\cite{Wainwright05}, and
compare it to both a baseline independence model, as well as a closely
related heuristic method that uses the ordinary sum-product (or belief
propagation) algorithm.

\subsection{Methods}

In Section~\ref{SecJointProcedure}, we described a generic procedure
for joint estimation and prediction.  Here we begin by describing the
special case of this procedure when the underlying variational method
is the tree-reweighted sum-product algorithm~\cite{Wainwright05}.  Any
instantiation of the tree-reweighted sum-product algorithm is
specified by a collection of edge weights $\rho_{st}$, one for each
edge $(s,t)$ of the graph.  The vector of edge weights must belong to
the spanning tree polytope; see Wainwright et al.~\cite{Wainwright05}
for further background on these weights and the reweighted algorithm.
Given a fixed set of edge weights $\rho$, the joint procedure based on
the tree-reweighted sum-product algorithm consists of the following
steps:
\begin{enumerate}
\item[1.] Given an initial set of i.i.d. data $\{X^1, \ldots,
X^\dnum\}$, we first compute the empirical marginal distributions
\begin{equation*}
\estim{\meanpar}_s(j) \defn \frac{1}{\dnum} \sum_{i=1}^\dnum \Ind[X^i_s
= j], \qquad \estim{\meanpar}_{st}(j,k) \defn \frac{1}{\dnum}
\sum_{i=1}^\dnum \Ind[X^i_s = j] \; \Ind[X^i_t =k],
\end{equation*}
and use them to compute the approximate parameter estimate
\begin{equation}
\label{EqnTRWParEst}
\eparest^\dnum_s(j) \defn \log \estim{\meanpar}_s(j), \qquad
\eparest^\dnum_s(j) \defn \rho_{st} \; \log
\frac{\estim{\meanpar}_{st}(j,k)}{\estim{\meanpar}_s(j)
\estim{\meanpar}_t(k)}.
\end{equation}
As shown in our previous work~\cite{Wainwright03a}, the
estimates~\eqref{EqnTRWParEst} are the global maxima of the surrogate
likelihood~\eqref{EqnSurLike} based on the convexified Bethe
approximation~\eqref{EqnDefnConvBethe} without any regularization term
(i.e., $\Reg = 0$).
\item[2.] Given the new noisy observation $Y$ of the
form~\eqref{EqnObsModel}, we incorporate it by by forming the new
exponential parameter
\begin{equation*}
\eparest^\dnum_s(\, \cdot \, ) + \datavec_s(\cdot; Y),
\end{equation*}
where equation~\eqref{EqnDefnEpy} defines $\datavec_s$ for the
Gaussian mixture model under consideration.

\item[3.]  We then compute approximate marginals $\taupar(\eparest +
\datavec)$ by running the TRW sum-product algorithm with edge
appearance weights $\rho_{st}$, using the message
updates~\eqref{EqnTRWMess}, on the graphical model distribution with
exponential parameter $\eparest + \datavec$.  We use the approximate
marginals (see equation~\eqref{EqnTRWMarg}) to construct the
prediction $\zapp$ in equation~\eqref{EqnAppPred}.
\end{enumerate}

We evaluated the tree-reweighted sum-product based on its increase in
mean-squared error (MSE) over the Bayes optimal
predictor~\eqref{EqnBayesOpt}.  Moreover, we compared the performance
of the tree-reweighted approach to the following alternatives:
\begin{enumerate}
\item[(a)] As a baseline, we used the \emph{independence model} in
which the mixture distributions at each node are all assumed to be
independent.  In this case, ML estimates of the parameters are given
by $\eparest_s(x_s) = \log \estim{\meanpar}_s(x_s)$, with all of the
coupling terms $\eparest_{st}(x_s, x_t)$ equal to zero.  The
prediction step reduces to computing the Bayes least squares estimate
at each node independently, based only on the local data $y_s$.
\item[(b)] The \emph{standard sum-product or belief propagation} (BP)
approach is closely related to the tree-reweighted sum-product method,
but based on the edge weights $\rho_{st} = 1$ for all edges.  In
particular, we first form the approximate parameter estimate
$\eparest$ using equation~\eqref{EqnTRWParEst} with $\rho_{st} = 1$.
As shown in our previous work~\cite{Wainwright03a}, this approximate
parameter estimate uniquely defines the Markov random field for which
the empirical marginals $\estim{\meanpar}_s$ and
$\estim{\meanpar}_{st}$ are fixed points of the ordinary belief
propagation algorithm.  We note that a parameter estimator of this
type has been used previously by other
researchers~\cite{Freeman00,Ross05}.  In the prediction step, we then
use the ordinary belief propagation algorithm (i.e., again with
$\rho_{st} = 1$) to compute approximate marginals of the graphical
model with parameter $\eparest + \datavec$.  Finally, based on these
approximate BP marginals, we compute the approximate predictor using
equation~\eqref{EqnAppPred}.
\end{enumerate}

\subsection{Comparisons}

\begin{figure}[h]
\begin{center}
\begin{tabular}{ccc}
\widgraph{0.35\textwidth}{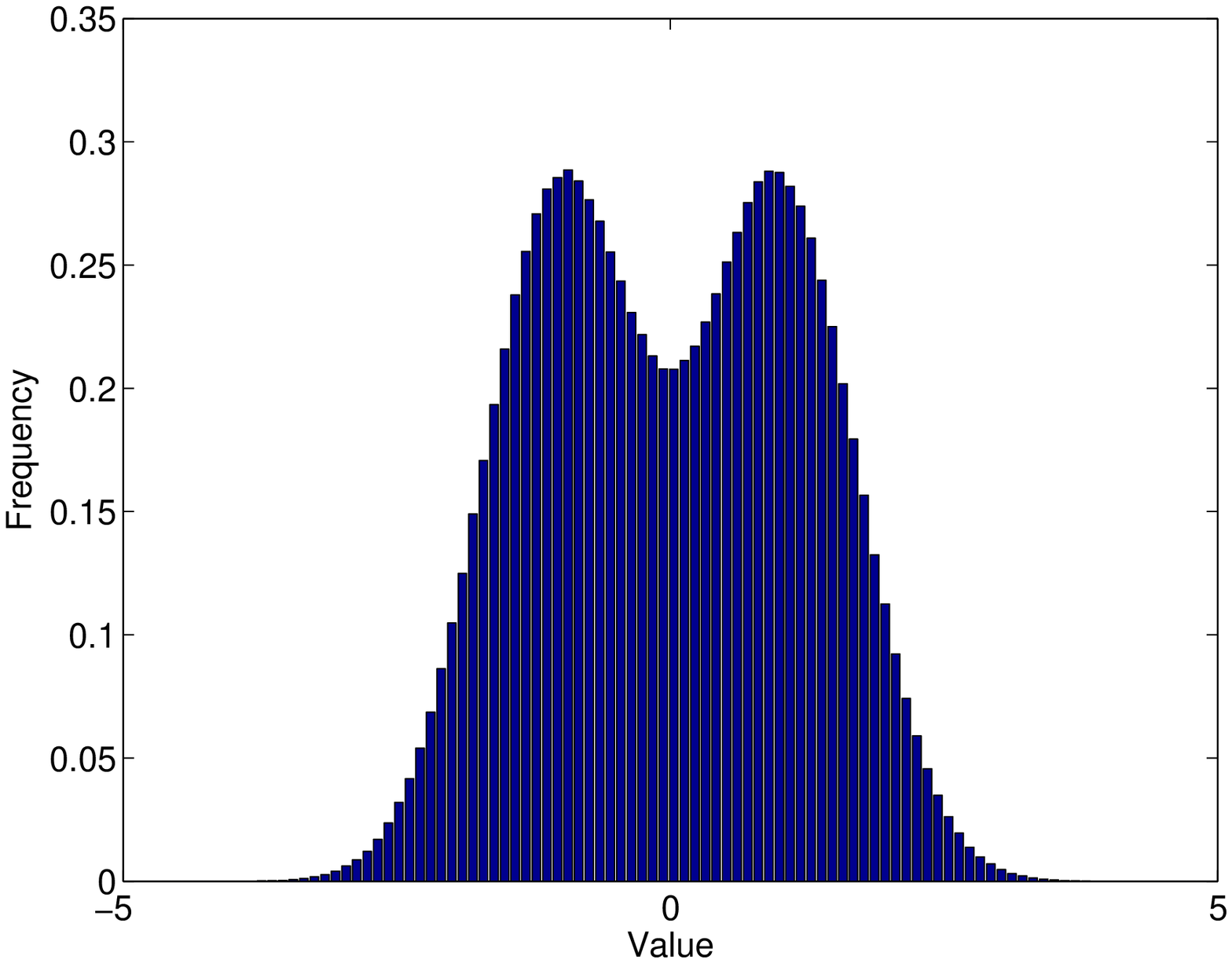} & &
\widgraph{0.35\textwidth}{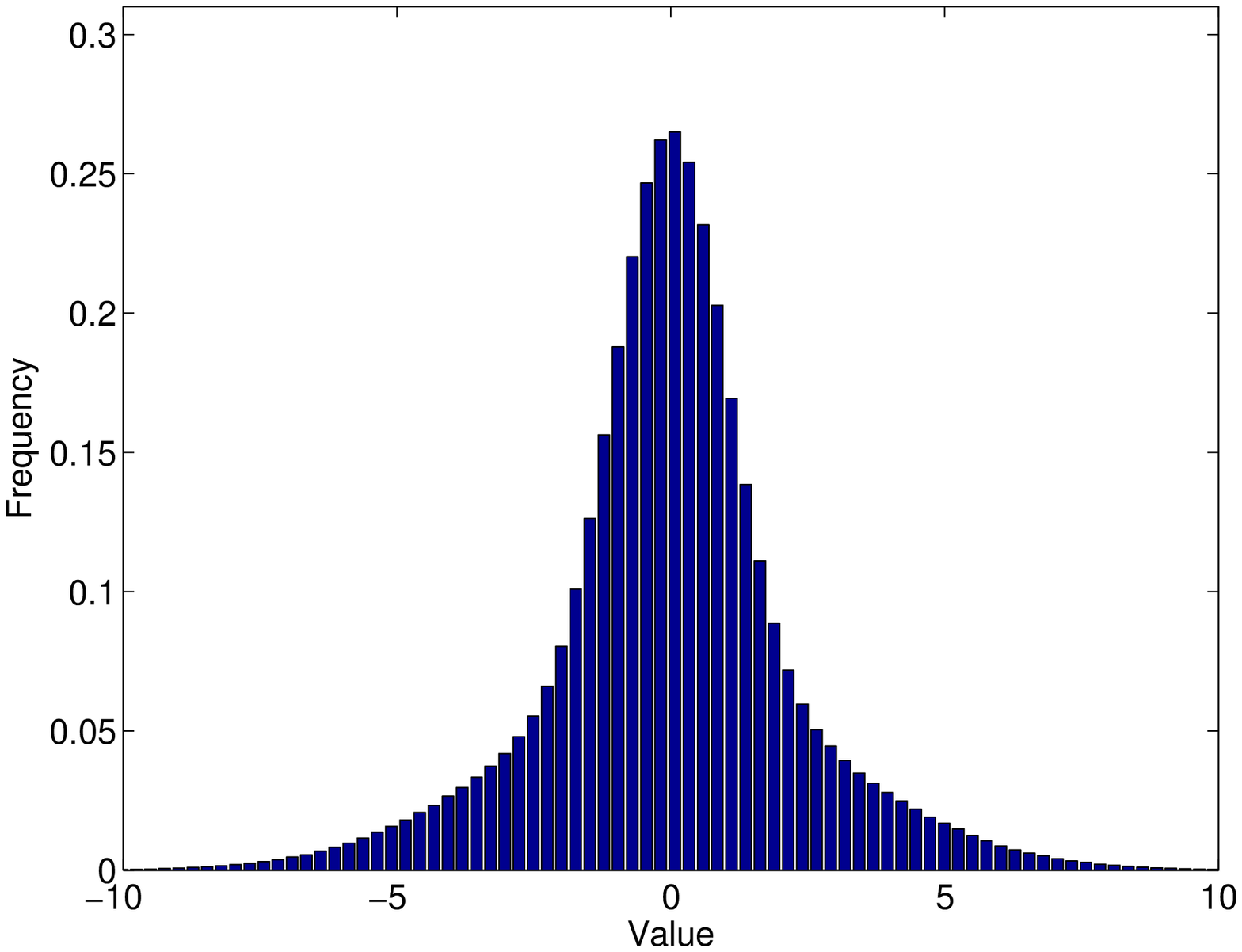} \\
(a) & & (b)
\end{tabular}
\end{center}
\caption{Histograms of different Gaussian mixture ensembles.  (a)
Ensemble A: a bimodal ensemble with $(\gmean_0, \gvar^2_0) = (-1,0.5)$
and $(\gmean_1, \gvar^2_1) = (1, 0.5)$.  (b) Ensemble B: mimics a
heavy-tailed distribution, with $(\gmean_0, \gvar^2_0) = (0, 1)$ and
$(\gmean_1, \gvar^2_1) = (0,9)$.}
\label{FigMixEnsemble}
\end{figure}

Although our methods are more generally applicable, here we show
representative results for $\statenum = 2$ components, and two
different types of Gaussian mixtures.
\begin{enumerate}
\item[(a)] Mixture ensemble A is bimodal, with components $(\gmean_0,
\gvar^2_0) = (-1,0.5)$ and $(\gmean_1, \gvar^2_1) = (1, 0.5)$.
\item[(b)] Mixture ensemble B was constructed with mean and variance
components $(\gmean_0, \gvar^2_0) = (0, 1)$ and $(\gmean_1, \gvar^2_1)
= (0,9)$; these choices serve to mimic heavy-tailed behavior.
\end{enumerate}
In both cases, each mixture component is equally weighted; see
Figure~\ref{FigMixEnsemble} for histograms of the resulting mixture ensembles.

Here we show results for a 2-D grid with $\vnum=64$ nodes.  Since the
mixture variables have $m = 2$ states, the coupling distribution can
be written as
\begin{equation*}
\pd{\eparams}{x} \propto \exp \big \{ \sum_{s \in \vertex} \eparam^*_s
x_s + \sum_{(s,t) \in \edge} \eparam^*_{st} x_s x_t \big \},
\end{equation*}
where $x \in \{-1, +1 \}^\vnum$ are ``spin'' variables indexing the
mixture components.  In all trials, we chose $\eparam^*_s = 0$ for all
nodes $s \in \vertex$, which ensures uniform marginal distributions
$\pd{\eparams}{x_s} = [0.5 \; \; 0.5]^T$ at each node.  We tested two
types of coupling in the underlying Markov random field:
\begin{enumerate}
\item[(a)] In the case of \emph{attractive coupling}, for each
coupling strength $\coupstr \in [0,1]$, we chose edge parameters as
$\eparam^*_{st} \sim \uni[0, \coupstr]$.
\item[(b)] In the case of \emph{mixed coupling}, for each coupling
strength $\coupstr \in [0,1]$, we chose edge parameters as
$\eparam^*_{st} \sim \uni[-\coupstr, \coupstr]$.
\end{enumerate}
Here $\uni[a,b]$ denotes a uniform distribution on the interval
$[a,b]$. In all cases, we varied the SNR parameter $\alsnr$, as
specified in the observation model~\eqref{EqnObsModel}, in the
interval $[0,1]$.

%%%%%%%%%%%%%%%%%%%%%%%%%%%%%%%%%%%%%%%%%%%%%%%%%%%%%%%%%%%%%%%%%%%%%%%%%%%%%%%%
Shown in Figure~\ref{FigMeshPlots} are 2-D surface plots of the
average percentage increase in MSE, taken over 100 trials, as a
function of the coupling strength $\coupstr \in [0,1]$ and the
observation SNR parameter $\alsnr \in [0,1]$ for the independence
model (left column), BP approach (middle column) and TRW method (right
column).  The top two rows show performance for attractive coupling,
for mixture ensemble A ((a) through (c)) and ensemble B ((d) through
(f)), whereas the bottom two row show performance for mixed coupling,
for mixture ensemble A ((g) through (i)) and ensemble B ((j) through
(l)).

\newcommand{\msize}{0.26}
\begin{figure*}[h!]
%\parbox{.96\textwidth}{ 
\bec
\begin{tabular}{ccc}
\widgraph{\msize\textwidth}{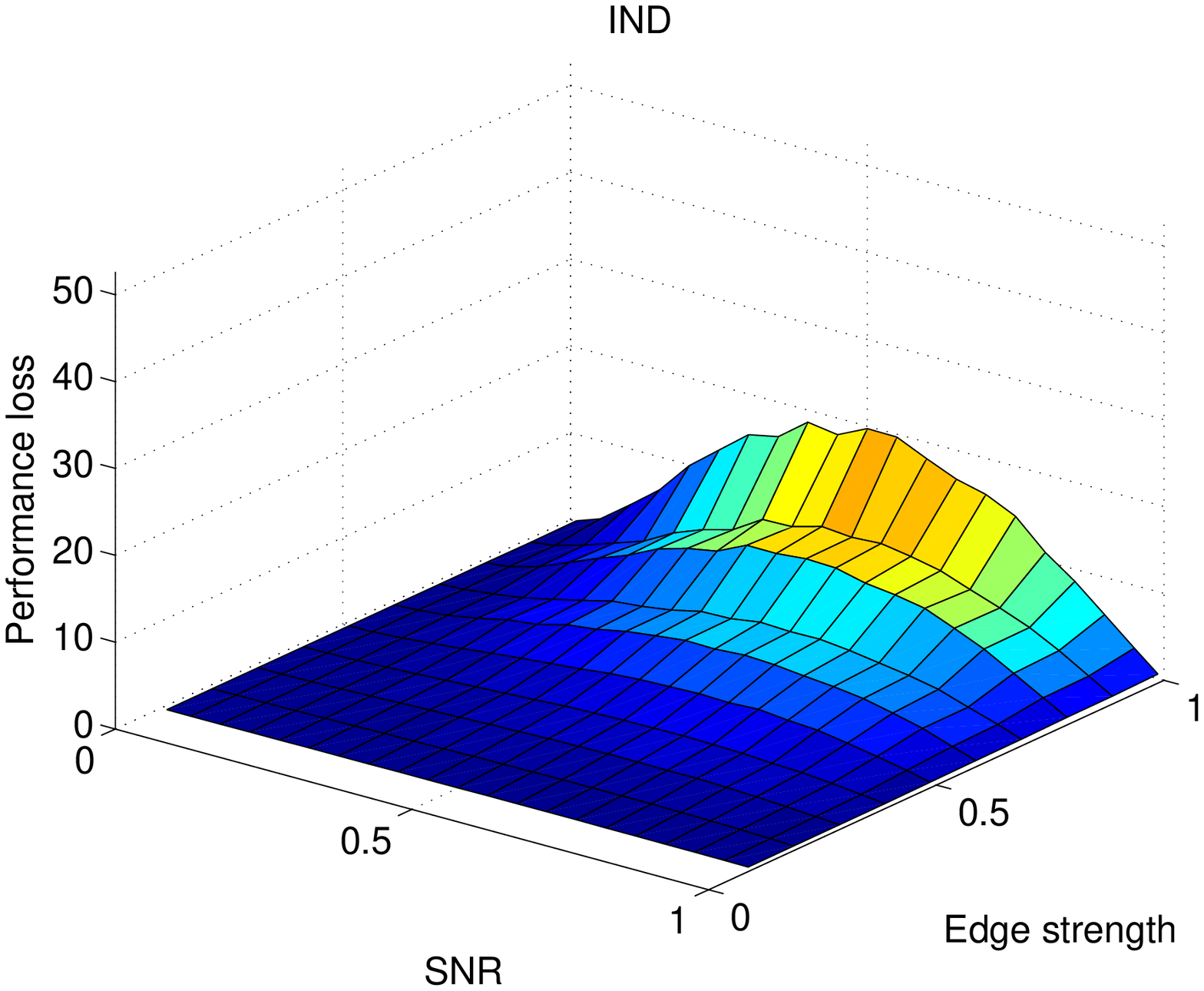} &
\widgraph{\msize\textwidth}{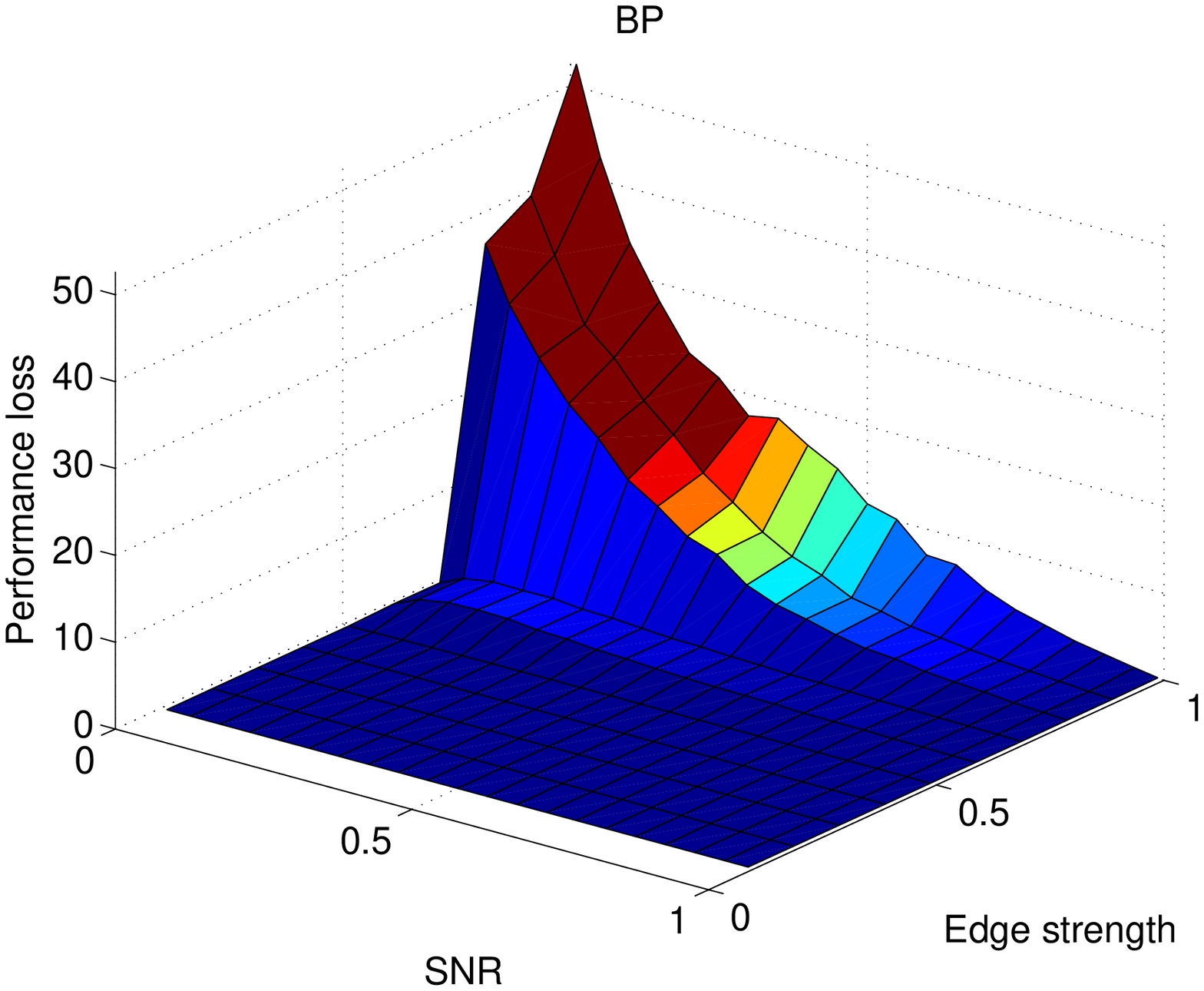} &
\widgraph{\msize\textwidth}{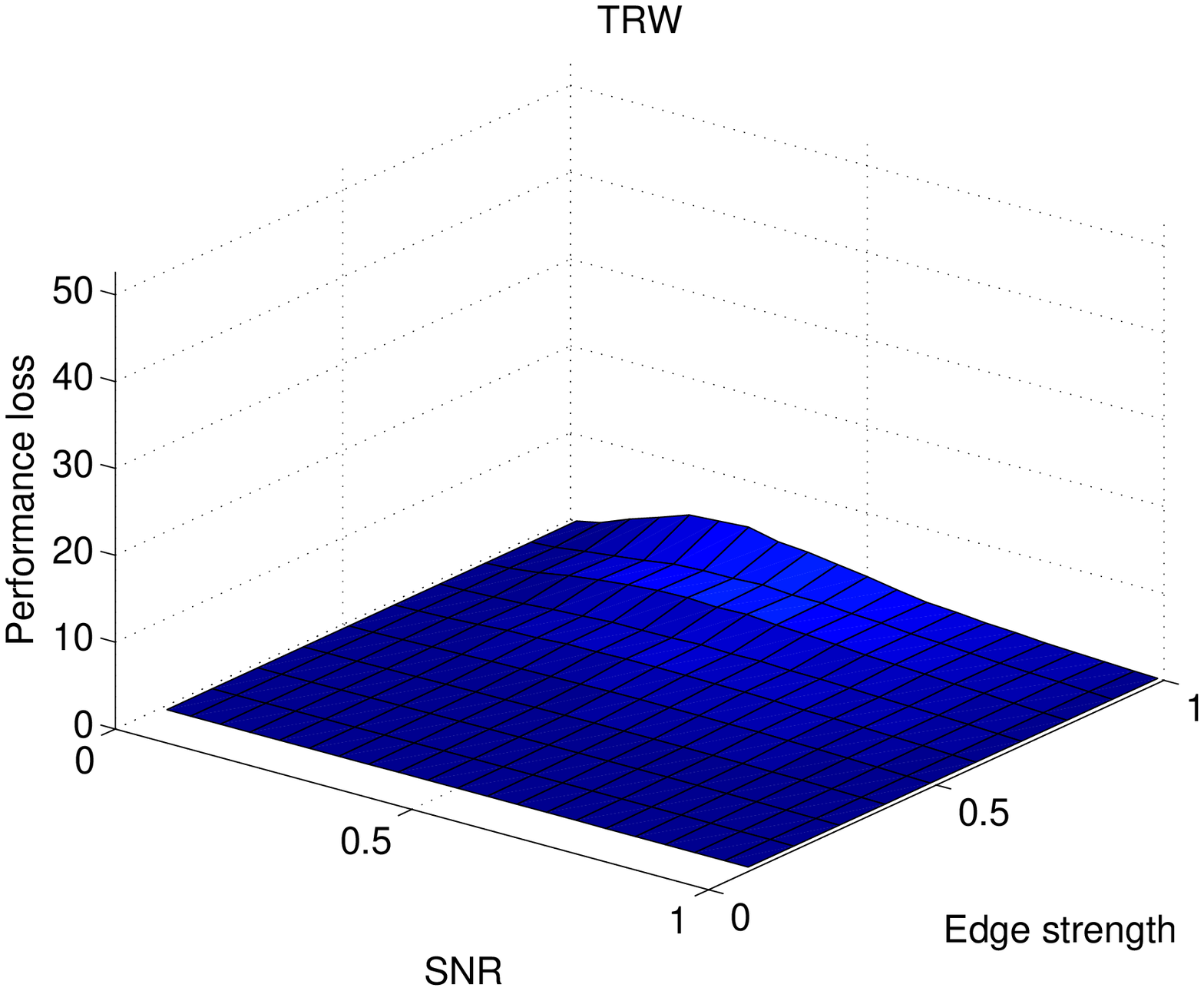} \\
(a) & (b) & (c) \\

\widgraph{\msize\textwidth}{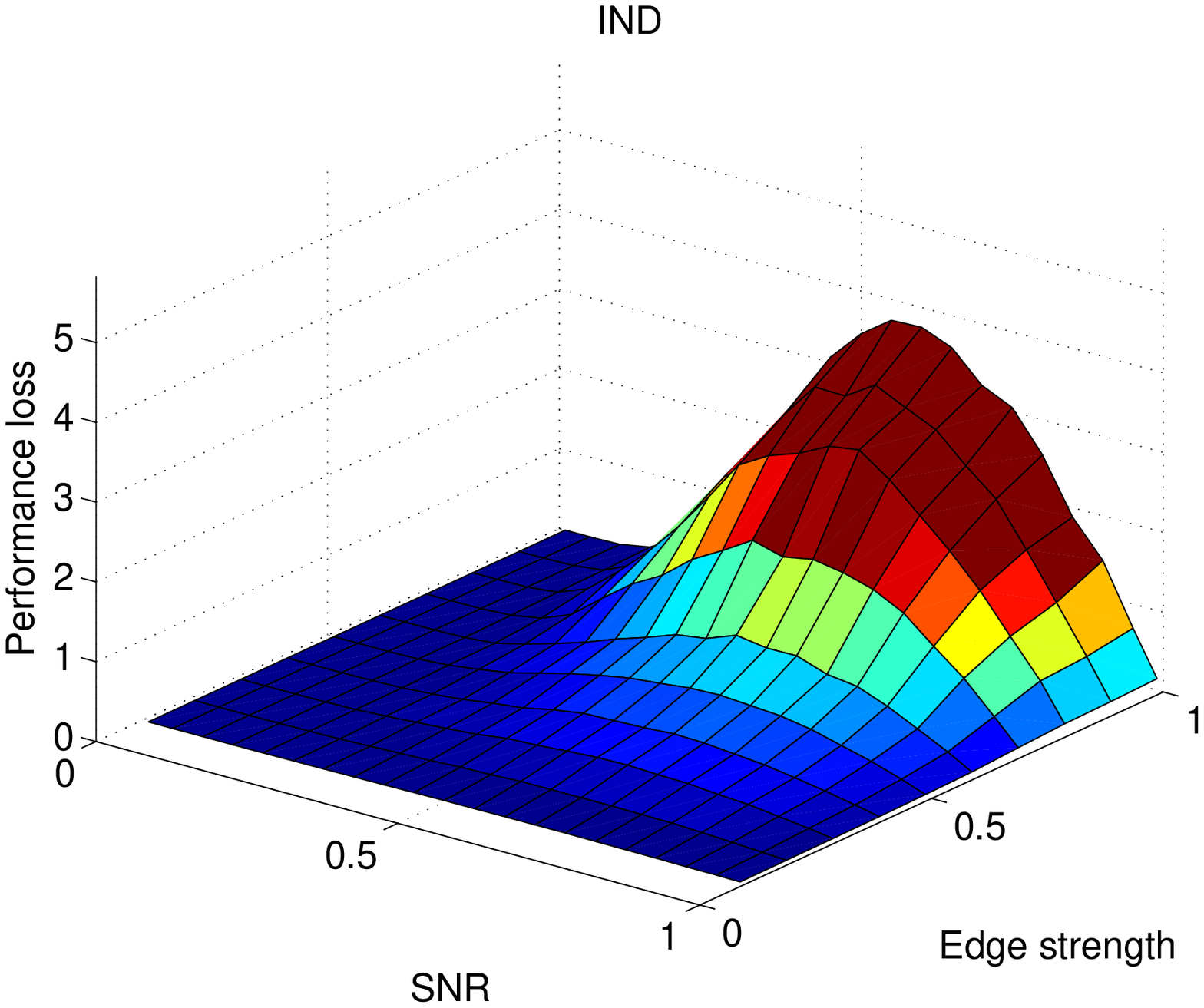} &
\widgraph{\msize\textwidth}{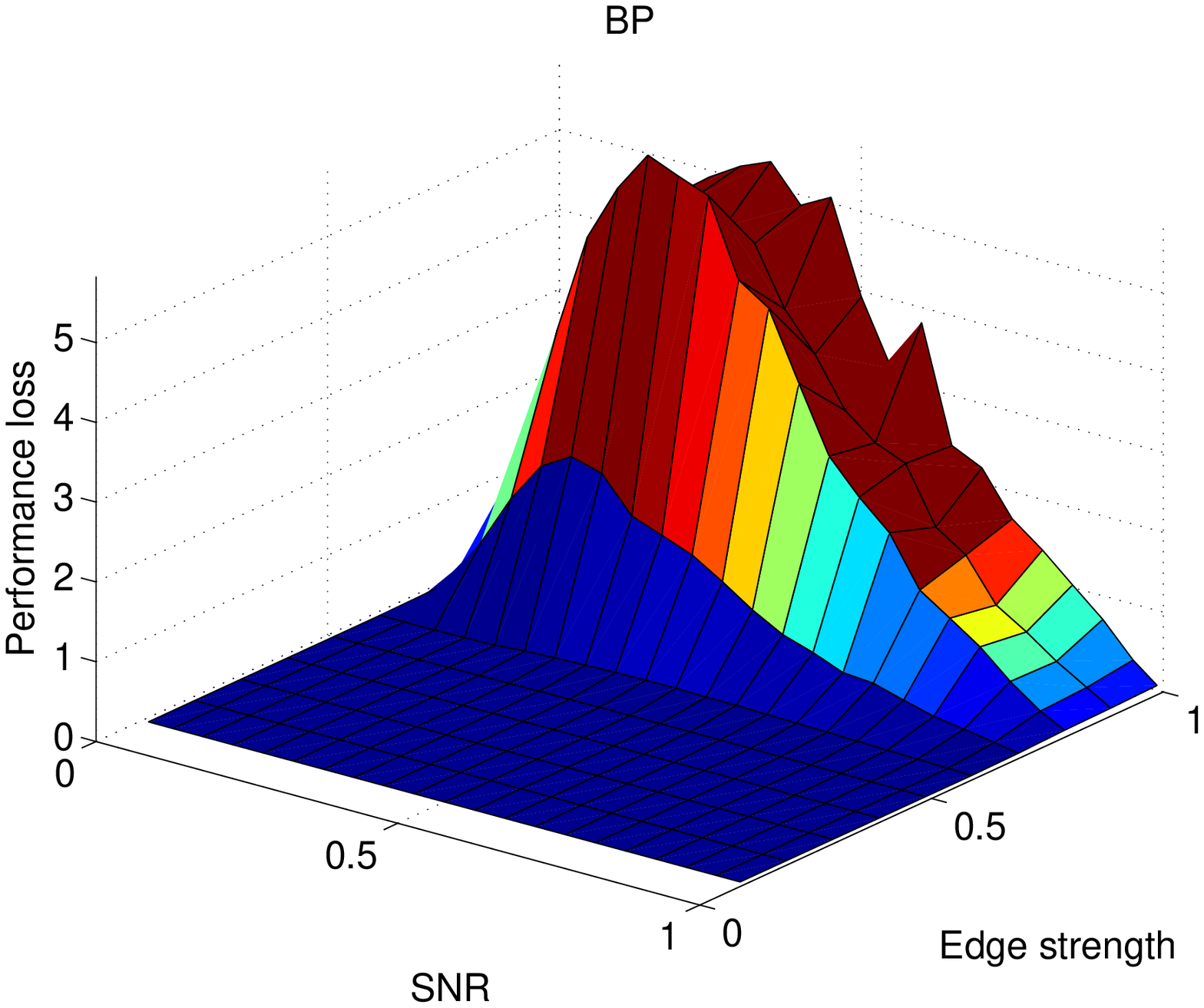} &
\widgraph{\msize\textwidth}{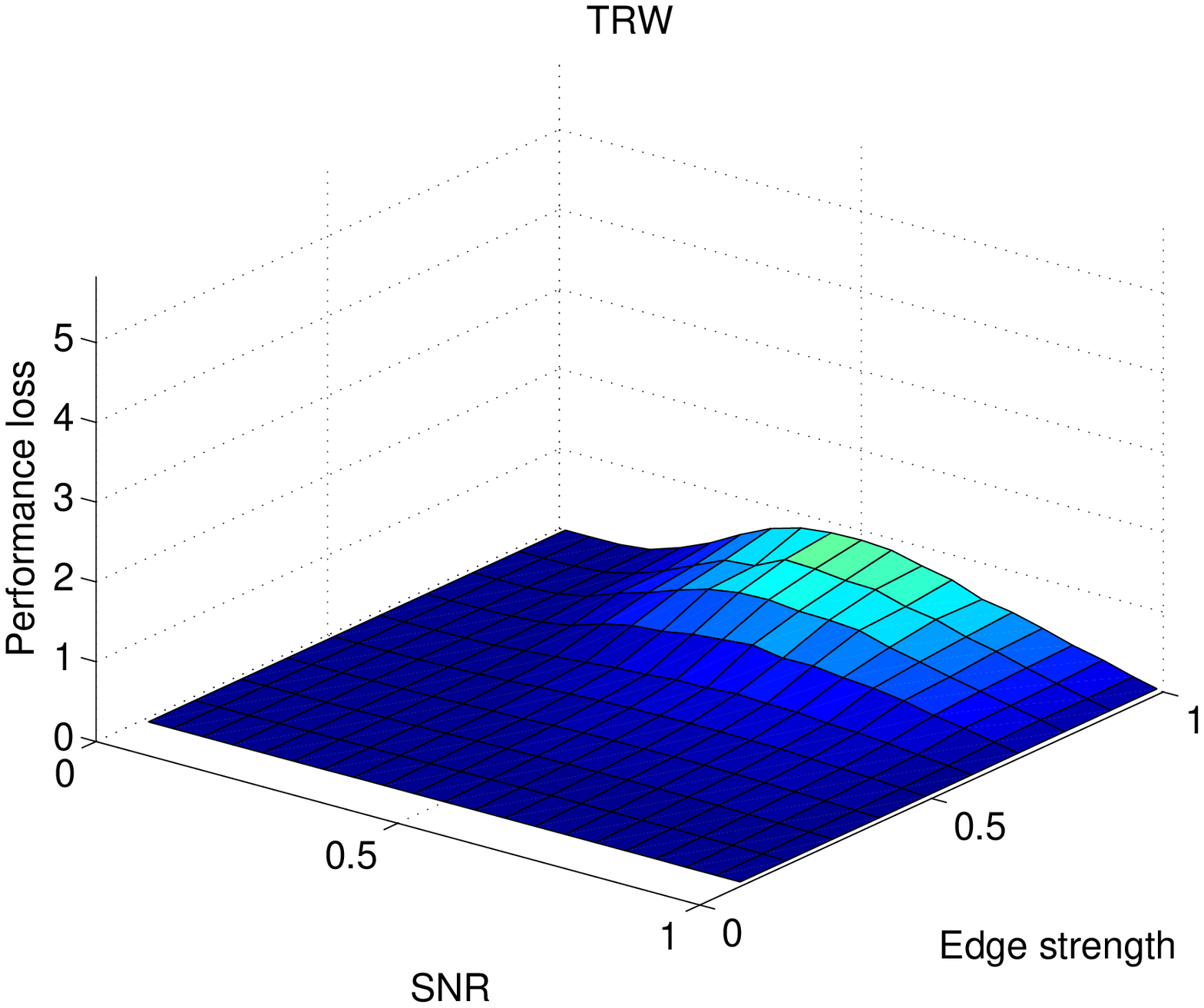} \\
(d) & (e) & (f) \\
\widgraph{\msize\textwidth}{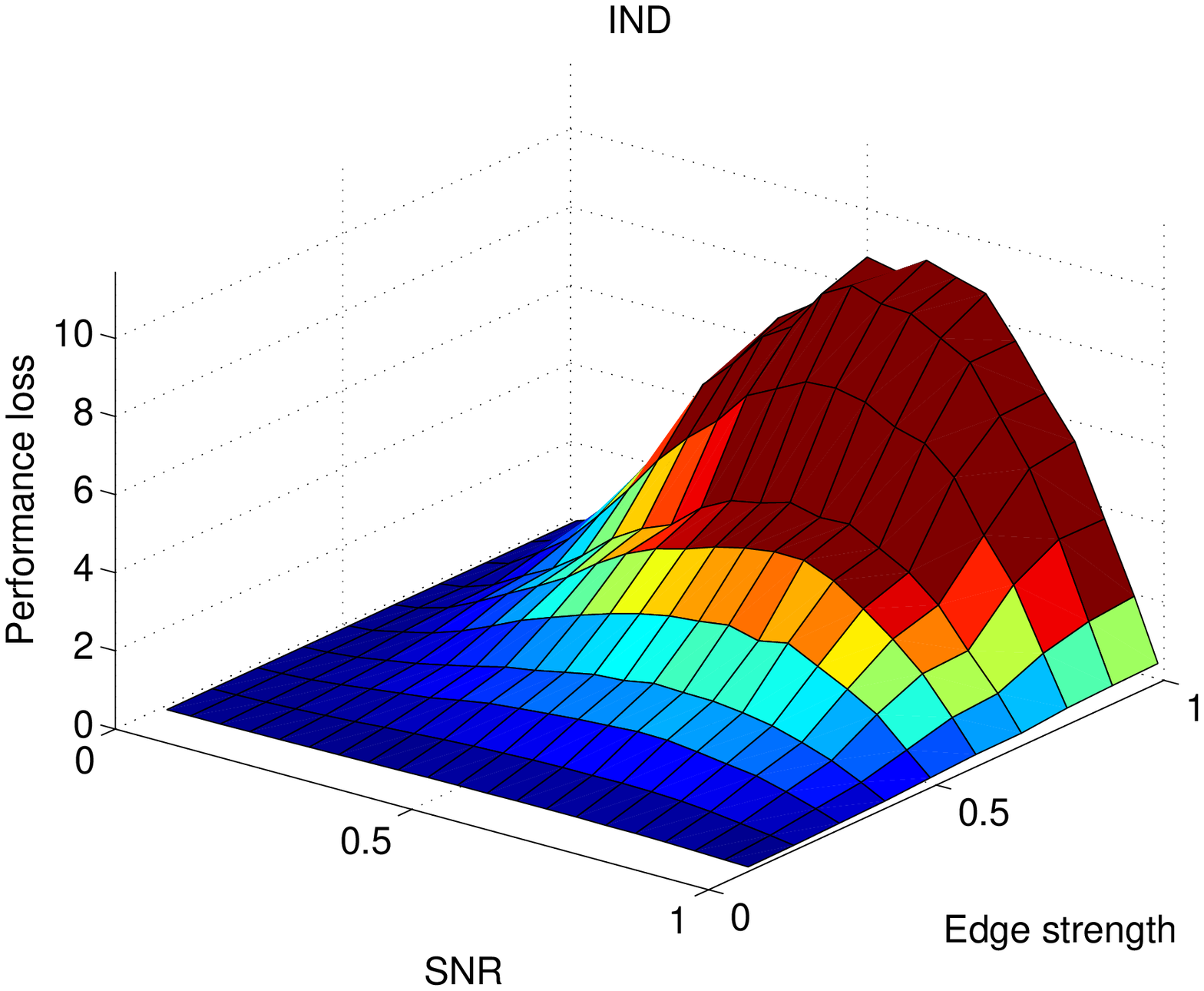} & 
\widgraph{\msize\textwidth}{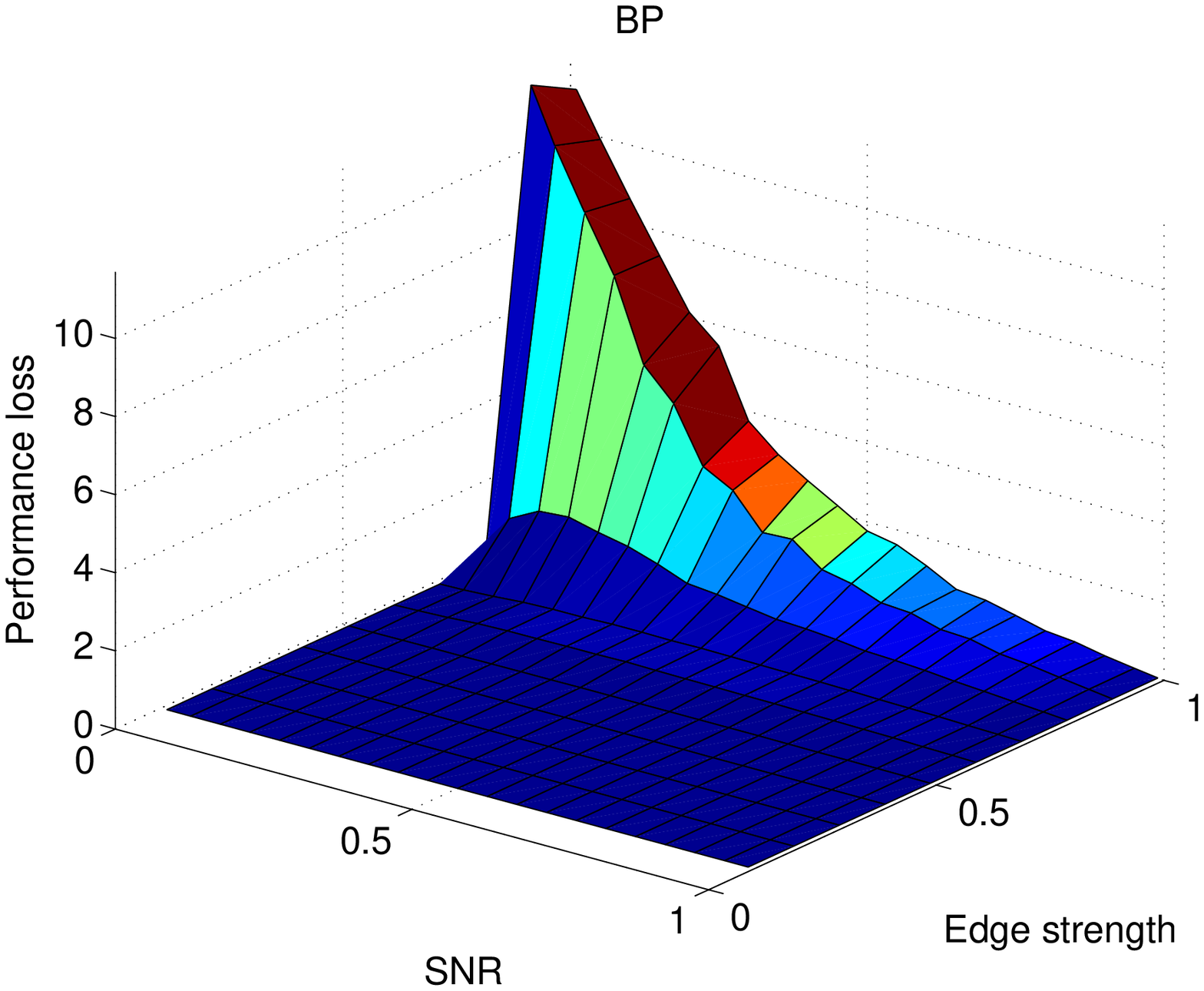} & 
\widgraph{\msize\textwidth}{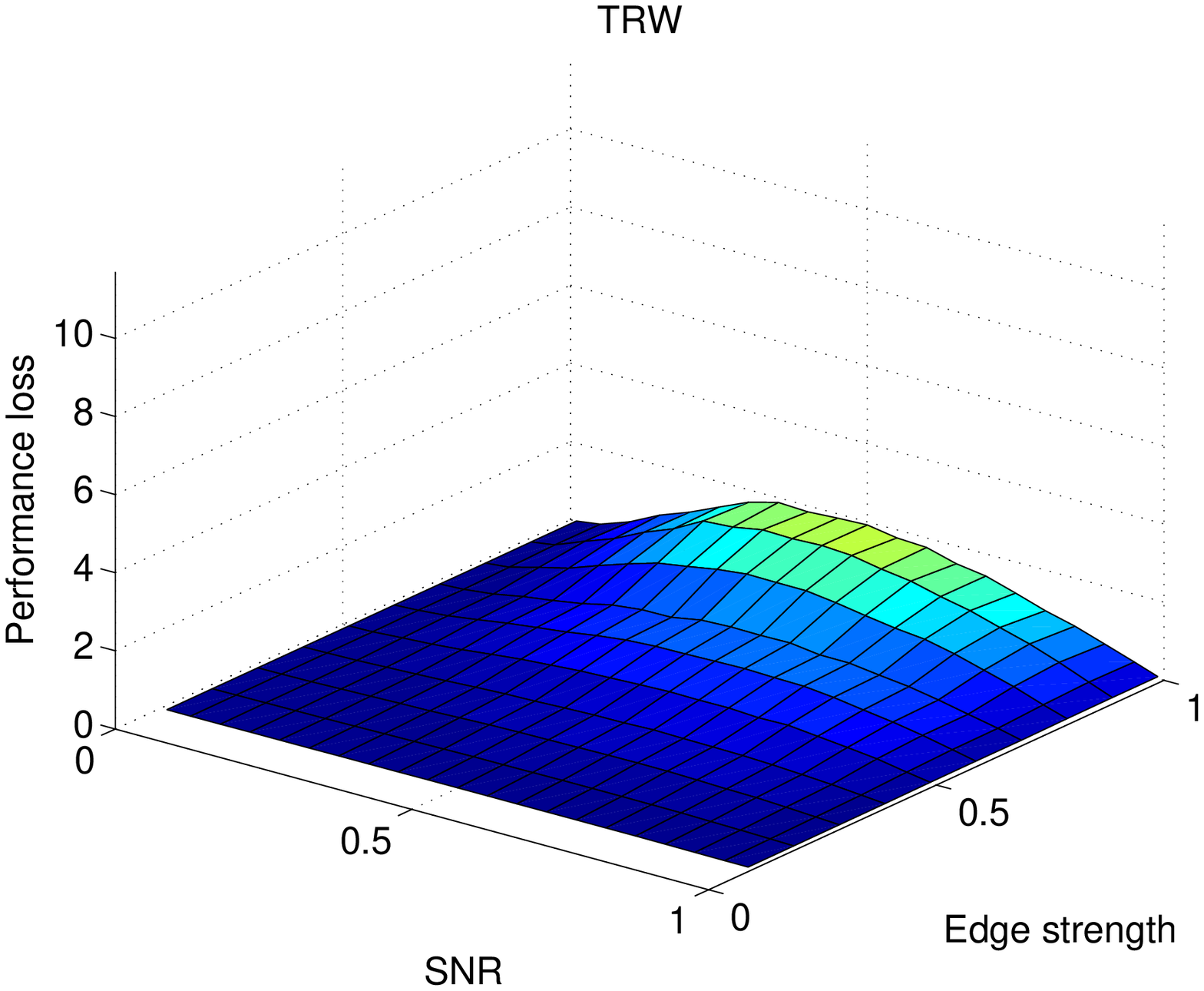} \\
(g) & (h) & (i)  \\
\widgraph{\msize\textwidth}{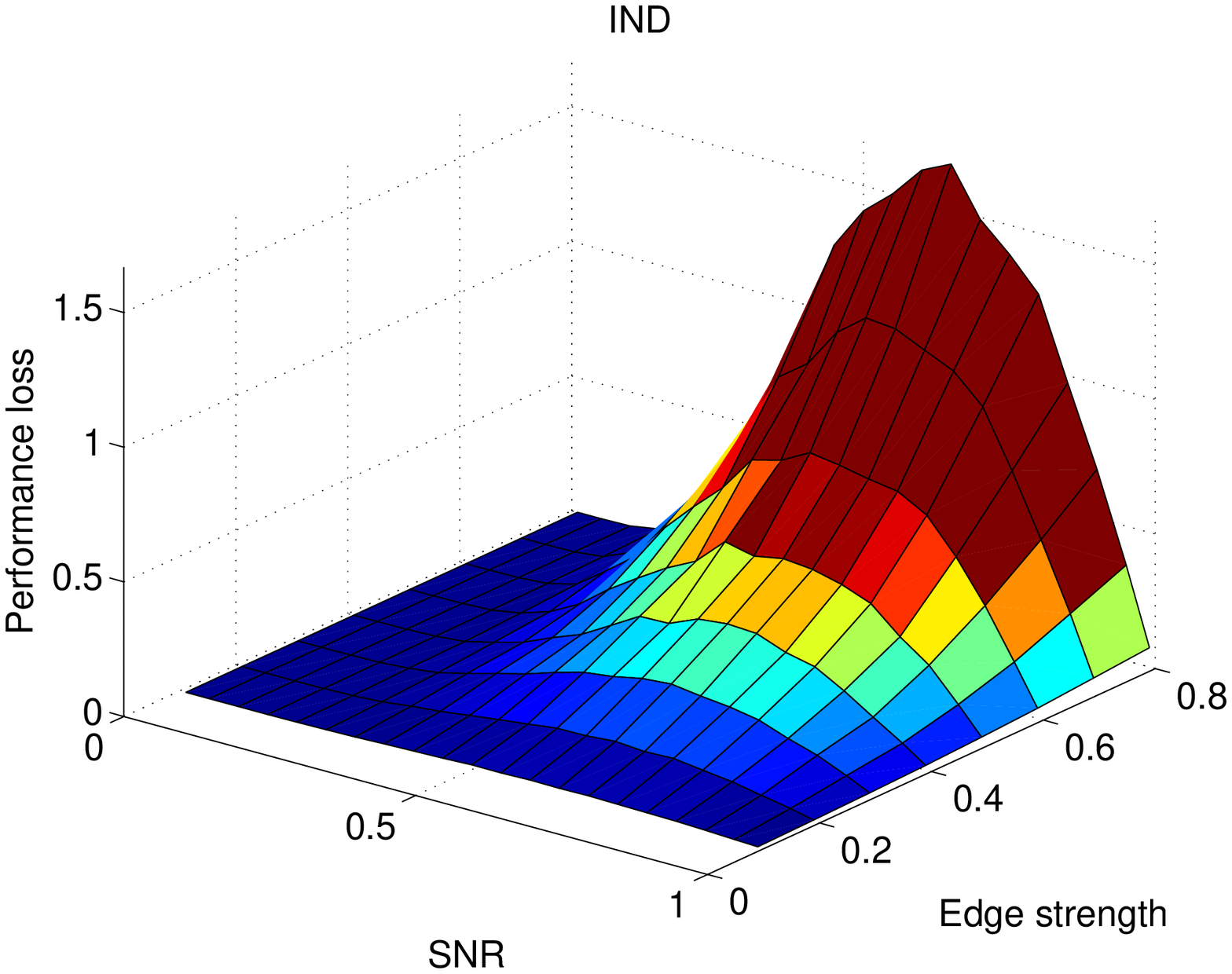} & 
\widgraph{\msize\textwidth}{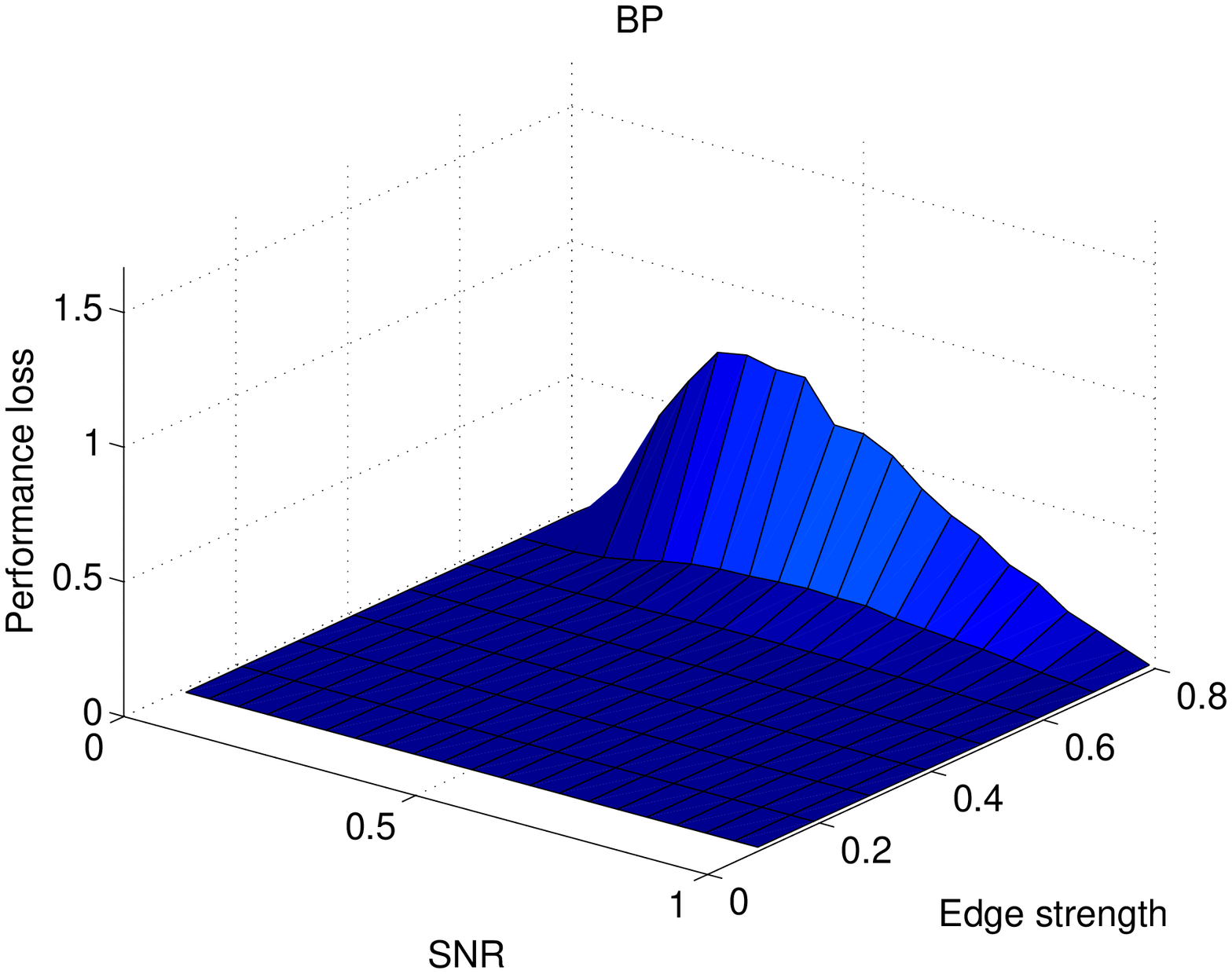} & 
\widgraph{\msize\textwidth}{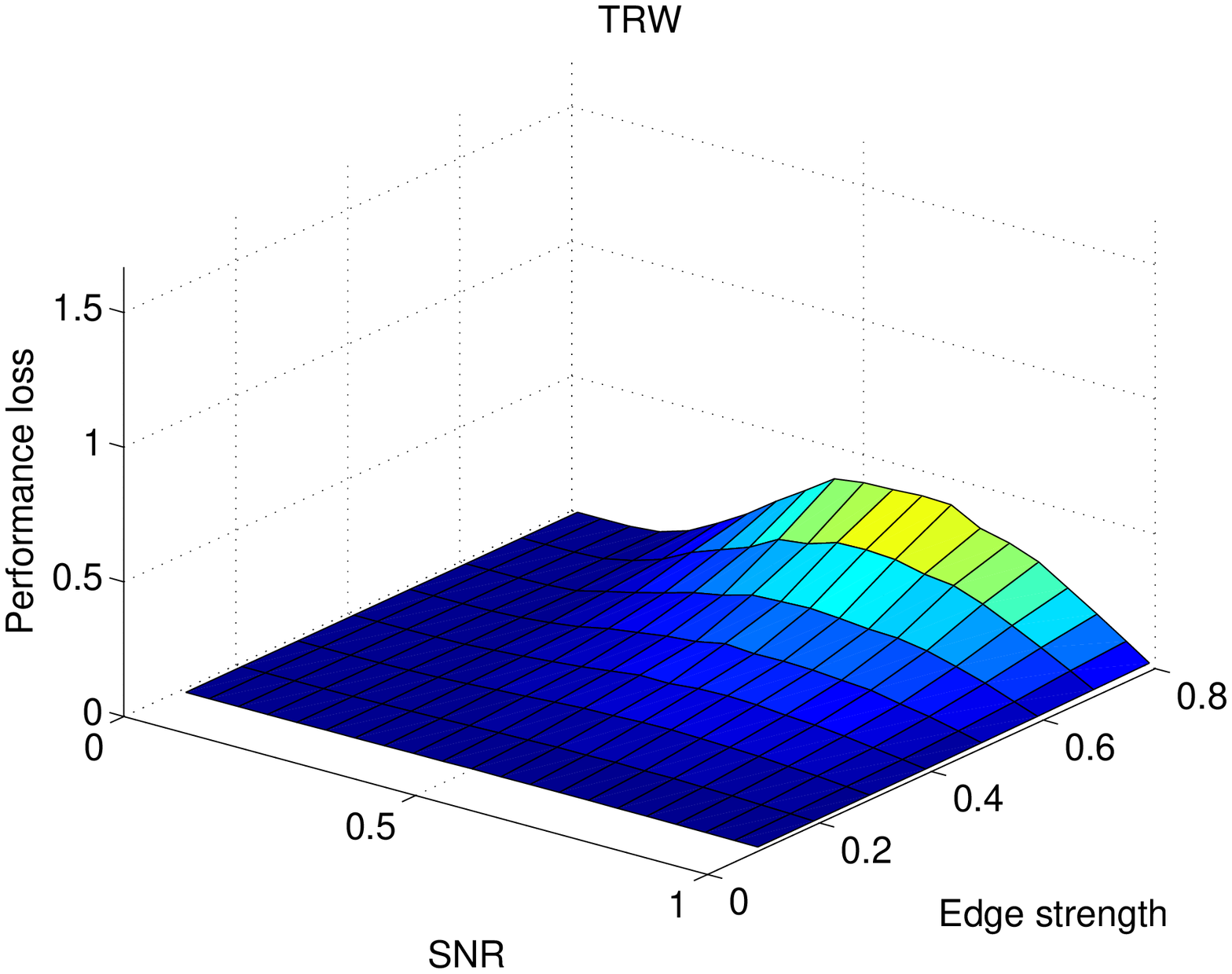} \\
(j) & (k) & (l)
\end{tabular}
\caption{Surface plots of the percentage increase in MSE relative to
Bayes optimum for different methods as a function of observation SNR
for grids with $\vnum = 64$ nodes.  Left column: independence model
(IND).  Center column: ordinary belief propagation (BP).  Right
column: tree-reweighted algorithm (TRW).  First row: Attractive
coupling and a Gaussian mixture with components $(\gmean_0, \gvar^2_0)
= (-1, 0.5)$ and $(\gmean_1, \gvar^2_1) = (1, 0.5)$.  Second row:
Attractive coupling and a Gaussian mixture with components $(\gmean_0,
\gvar^2_0) = (0,1)$ and $(\gmean_0, \gvar^2_1) = (0,9)$.  Third row:
Mixed coupling and a Gaussian mixture with components $(\gmean_0,
\gvar^2_0) = (-1, 0.5)$ and $(\gmean_1, \gvar^2_1) = (1, 0.5)$.
Fourth row: Mixed coupling and a Gaussian mixture with components
$(\gmean_0, \gvar^2_0) = (0,1)$ and $(\gmean_0, \gvar^2_1) = (0,9)$.
}
\label{FigMeshPlots}
\enc
%}
\end{figure*}

First, observe that for weakly coupled problems ($\coupstr \approx
0$), whether attractive or mixed coupling, all three
methods---including the independence model---perform quite well, as
should be expected given the weak dependency between different nodes
in the Markov random field.  Although not clear in these plots, the
standard BP method outperforms the TRW-based method for weak coupling;
however, both methods lose than than 1\% in this regime.  As the
coupling is increased, the BP method eventually deteriorates quite
seriously; indeed, for large enough coupling and low/intermediate SNR,
its performance can be worse than the independence (IND) model.  This
deterioration is particularly severe for the case of mixture ensemble
A with attractive coupling, where the percentage loss in BP can be as
high as 50\%.  Looking at alternative models (in which phase
transitions are known), we have found that this type of rapid
degradation coincides with the appearance of multiple fixed points.
In contrast, the behavior of the TRW method is extremely stable, which
is consistent with our theoretical results.

%%%%%%%%%%%%%%%%%%

\subsection{Comparison between theory and practice}

We now compare the practical behavior of the tree-reweighted
sum-product algorithm to the theoretical predictions from
Theorem~\ref{ThmBound}.  In general, we have found that in
quantitative terms, the bounds~\eqref{EqnKeyBound} are rather
conservative---in particular, the TRW sum-product method performs much
better than the bounds would predict.  However, here we show how the
bounds can capture qualitative aspects of the MSE increase in
different regimes.

\begin{figure}[h]
\begin{center}
\begin{tabular}{cc}
\widgraph{.45\textwidth}{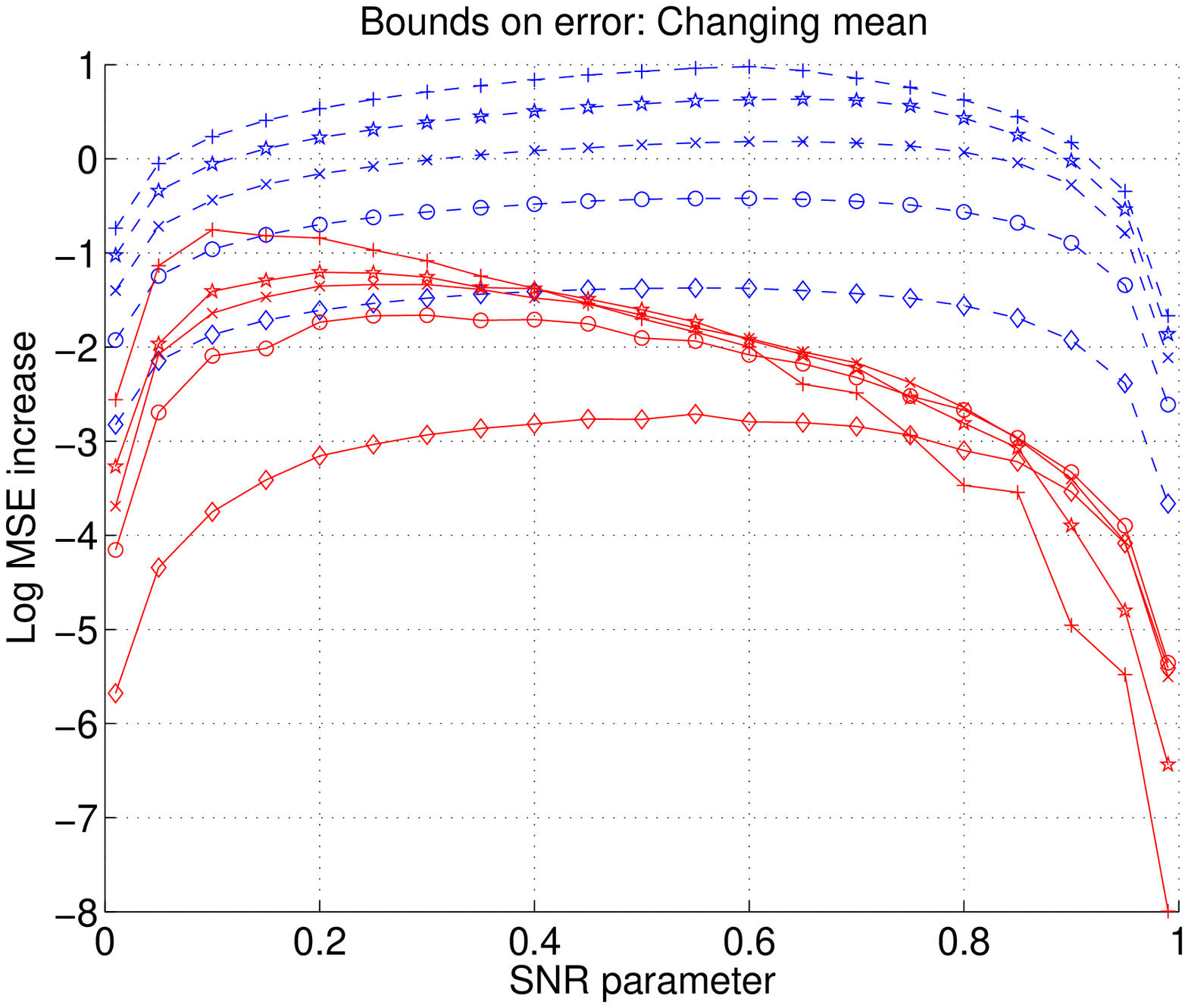} &
\widgraph{.45\textwidth}{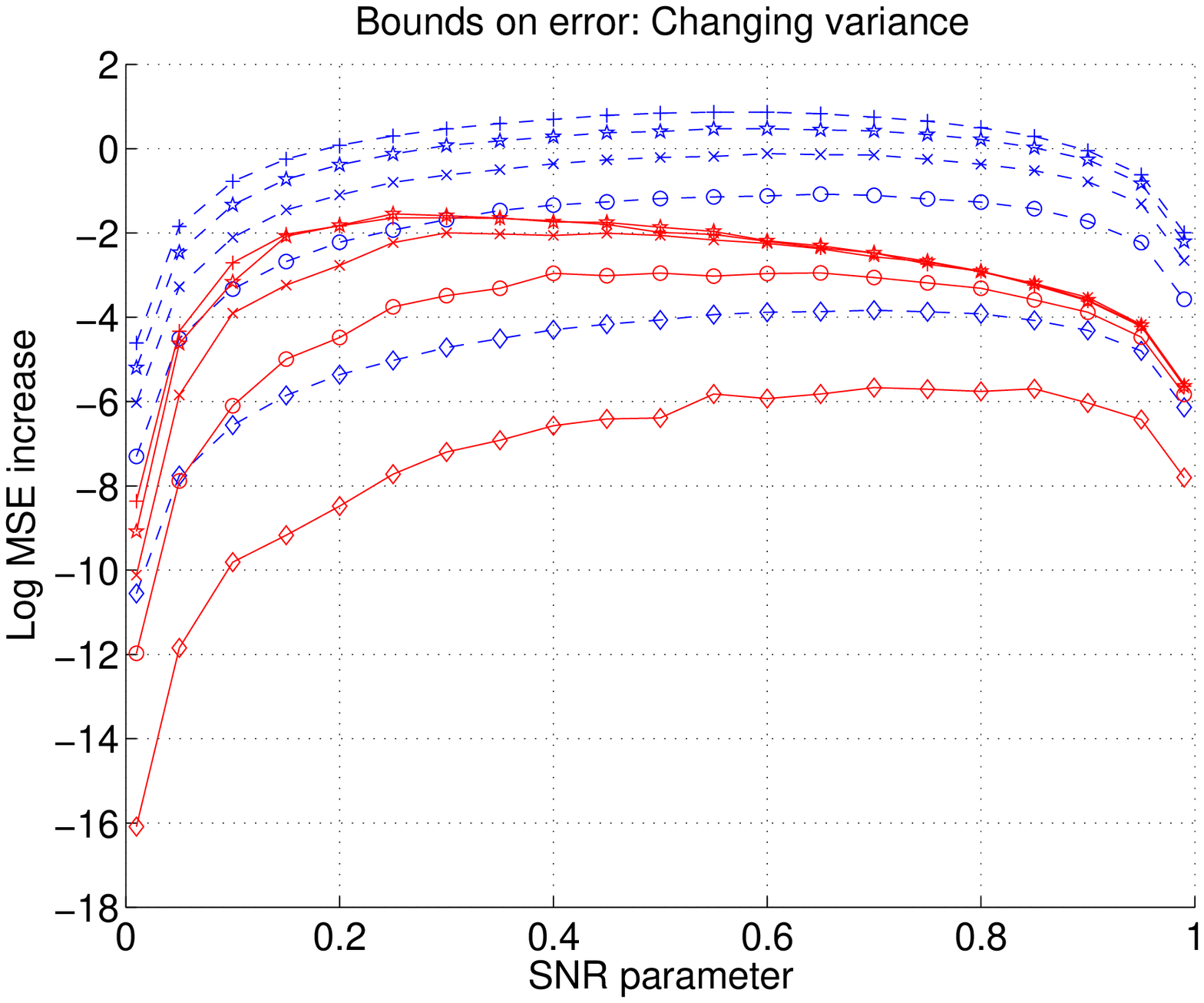} \\
(a) & (b)
\end{tabular}
\end{center}
\caption{Comparison of actual MSE increase and upper bounds for grid
with $\vnum = 64$ nodes with attractive coupling.  (a) Equal variances
$\sigma_0^2 = \sigma_1^2 = 0.5$, and mean vectors $(\gmean_0,
\gmean_1)$ ranging from $(-0.5, 0.5)$ to $(-2.5, 2.5)$.  (b) Equal
mean vectors $\gmean_0 = \gmean_1 = 0$, and variances $(\sigma_0^2,
\sigma_1^2)$ ranging from $(1, 1.25)$ to $(1, 25)$. }
\label{FigCompBounds}
\end{figure}

Figure~\ref{FigCompBounds} provides plots of the actual MSE increase
for the TRW algorithm (dotted blue lines), compared to the theoretical
bound~\eqref{EqnKeyBound} (solid red lines), for the grid with $\vnum
= 64$ nodes, and attractive coupling of strength $\coupstr = 0.70$.
For all comparisons in both panels, we used $\Lcon = 0.10$, which
numerical calculations showed to be a reasonable choice for this
coupling strength.  (Overall, changes in the constant $\Lcon$
primarily cause the bounds to shift up and down on the log scale, and
so do not overly affect the qualitative comparisons given here.)
Panel (a) provides the comparison ensembles of type A, with fixed
variances $\sigma^2_0 = \sigma^2_1 = 0.5$ and mean vectors $(\gmean_0,
\gmean_1)$ ranging from $(-0.5, 0.5)$ to $(-2.5, 2.5)$.  Note how the
bounds capture the qualitative behavior for low SNR, for which the
difficulty of the problem increases as the mean separation is
increased.  In contrast, in the high SNR regime, the bounds are
extremely conservative, and fail to predict that the sharp drop-off in
error as the SNR parameter $\alsnr$ approaches one.  This drop-off is
particularly pronounced for the ensemble with largest mean separation
(marked with $+$).  Panel (b) provides a similar comparison for
ensembles of type B, with fixed mean vectors $\gmean_0 = \gmean_1 =
0$, and variances $(\sigma^1_0, \sigma^2_1)$ ranging from $(1, 1.25)$
to $(1, 25)$.  In this case, although the bounds are still very
conservative in quantitative terms, they reasonably capture the
qualitative behavior of the error over the full range of SNR.

\section{Discussion}
\label{SecDiscussion}

Key challenges in the application of Markov random fields include the
estimation (learning) of model parameters, and performing prediction
using noisy samples (e.g., smoothing, interpolation, denoising).  Both
of these problems present substantial computational challenges for
general Markov random fields.  In this paper, we have described and
analyzed methods for joint estimation and prediction that are based on
convex variational methods.  Our central result is that using
inconsistent parameter estimators can be beneficial in the
computation-limited setting.  Indeed, our results provide rigorous
confirmation of the fact that using parameter estimates that are
``systematically incorrect'' is helpful in offsetting the error
introduced by using an approximate method during the prediction step.
In concrete terms, we demonstrated that a joint prediction/estimation
method using the tree-reweighted sum-product algorithm yields good
performance across a wide range of experimental conditions.  Although
our work has focused on a particular scenario, we suspect that similar
ideas and techniques will be useful in related applications of
approximate methods for learning and prediction.

\subsection*{Acknowledgments}
This work was supported by an Alfred P. Sloan Foundation Fellowship,
an Okawa Foundation Research Fellowship, an Intel Corporation
Equipment Grant, and NSF Grant DMS-0528488.

\appendix

\section{Tree-based relaxation}
\label{AppStrict}
As an illustration on the single cycle on $3$ vertices, the
pseudomarginal vector with elements
\begin{equation*}
\taupar_s(x_s) = \begin{bmatrix} 0.5 \\ 0.5 \end{bmatrix} \; \;
\mbox{for $s = 1,2,3$} \quad \mbox{and} \quad \taupar_{st}(x_s, x_t) =
\begin{bmatrix} \alpha_{st} & 0.5 - \alpha_{st} \\ 0.5 - \alpha_{st} &
\alpha_{st}
 \end{bmatrix} 
\end{equation*}
belongs to $\myLocset(\graph)$ for all choices $\alpha_{st} \in [0,
  0.5]$, but fails to belong to $\myMargset(\graph)$, for instance,
when $\alpha_{12} = \alpha_{23} = \alpha_{13} = 0$.

\section{Proof of Lemma~\ref{LemExact}}
\label{AppExact}

Using Lemma~\ref{LemBasic} and the mean value theorem, we write
\begin{eqnarray*}
\meanpar(\eparam + \delta) - \meanpar(\eparam) & = & \nabla
\Partc(\eparam + \delta) - \nabla \Partc(\eparam) \\
& = & \nabla^2 \Partc(\eparam + t \delta) \delta
\end{eqnarray*}
for some $t \in (0,1)$.  Hence, it suffices to show that the
eigenspectrum of the Hessian \mbox{$\nabla^2 \Partc(\eparam) =
\cov_\eparam \{ \clipot(X) \}$} is uniformly bounded above by $L <
+\infty$.  The functions $\clipot$ are all 0-1 valued indicator
functions, so that the diagonal elements of $\cov_\eparam \big\{
\clipot(X) \big \}$ are bounded above---in particular,
$\var(\clipot_\sumind(X)) \leq \frac{1}{4}$ for any index $\sumind \in
\{1, \ldots, \df \}$.  Consequently, we have
\begin{eqnarray*}
\lambda_{\operatorname{max}} (\cov_\eparam \{\clipot(X) \}) & \leq &
\sum_{\alpha=1}^\df \lambda_\alpha(\cov_\eparam \{ \clipot(X) \} \; =
\; \trace(\cov_\eparam\{ \clipot(X) \} \; = \; \frac{\df}{4}
\end{eqnarray*}
as required.

%%%%%%%%%%%%%%%%%%%%%%%%%%%%%%%%%%%%%%%%%%%%%%%%%%%%%%%%%%%%%%%%%%%%%%

\section{Proof of Lemma~\ref{LemConv}}
\label{AppConv}
Consider a spanning tree $\treegr$ of $\graph$ with edge set
$\edge(\treegr)$.  Given a vector $\taupar \in \myLocset(\graph)$, we
associate with $\treegr$ a subvector $\tauparr{\treegr}$ formed by
those components of $\taupar$ associated with vertices $\vertex$ and
edges $\edge(\treegr)$.  Note that by construction $\tauparr{\treegr}
\in \myLocset(\treegr) = \myMargset(\treegr)$.  The mapping $\taupar
\mapsto \tauparr{\treegr}$ can be represented by a projection matrix
$\Projt \in \real^{\df(\treegr) \times \df}$ with the block structure
\[
\Projt \defn \begin{bmatrix} I_{\df(\treegr) \times \df(\treegr)} &
0_{\df(\treegr) \times (\df-\df(\treegr))} \end{bmatrix}.
\]
In this definition, we are assuming for convenience that $\taupar$ is
ordered such that the $\df(\treegr)$ components corresponding to the
tree $\treegr$ are placed first.  With this notation, we have $\Projt
\taupar =
\begin{bmatrix} \tauparr{\treegr} & 0
\end{bmatrix}'$.

By our construction of the function $\Cbsur$, there exists a
probability distribution \mbox{$\edgewvec \defn \{ \edgew(\treegr) \;
| \; \treegr \in \tract \}$} such that $\Cbsur(\taupar) =
\sum_{\treegr \in \tract} \edgew(\treegr) \Dualc(\tauparr{\treegr})$,
where $\Dualc(\tauparr{\treegr})$ denotes the negative entropy of the
tree-structured distribution defined by the vector of marginals
$\tauparr{\treegr}$.  Hence, the Hessian of $\Cbsur$ has the
decomposition
\begin{eqnarray}
\label{EqnDecomp}
\nabla^2 \Cbsur(\taupar) & = & \sum_{\treegr \in \tract}
\edgew(\treegr) (\Projt)' \, \nabla^2 \Dualc(\tauparr{\treegr})
(\Projt)
\end{eqnarray}
To check dimensions of the various quantities, note that $\nabla^2
\Dualc(\tauparr{\treegr})$ is a $\df(\treegr) \times \df(\treegr)$
matrix, and recall that each matrix $\Projt \in \real^{\df(\treegr)
\times \df}$.

Now by Lemma~\ref{LemExact}, the eigenvalues of the $\nabla^2 \Partc$
are uniformly bounded above; hence, the eigenvalues of $\nabla^2
\Dualc$ are uniformly bounded away from zero.  Hence, for each tree
$\treegr$, there exists a constant $\Ccon_T$ such that for all $z \in
\real^\df$
\[
z' (\Projt)' \, \nabla^2 \Dualc(\tauparr{\treegr}) (\Projt) z \; \geq
\; \Ccon_T \|\Projt z \|^2 \; = \; \Ccon_T \|z\{\treegr\}\|^2.
\]
Substituting this relation into our decomposition~\eqref{EqnDecomp}
and expanding the sum over $\treegr$ yields
\begin{eqnarray}
z' \nabla^2 \Cbsur(\taupar) z & \geq & \sum_{\treegr \in \tract}
\edgew(\treegr) \Ccon_T \|z\{\treegr\}\|^2 \nonumber \\ 
\label{EqnTmpLower}
& = & \big[\sum_{\treegr \in \tract} \edgew(\treegr)
\Ccon_\treegr\big] \sum_{s \in \vertex} \|z\{s\}\|^2 + \sum_{(s,t) \in
\edge} \big[ \sum_{\treegr \in \tract} \edgew(\treegr) \Ccon_\treegr
\Ind[(s,t) \in \edge(\treegr)] \; \big] \; \|z\{(s,t) \} \|^2. \qquad
\end{eqnarray}
Defining $\Ccon^* \defn \min_{\treegr \in \tract} \Ccon_\treegr$, we have the
lower bounds
\begin{eqnarray*}
\big[\sum_{\treegr \in \tract} \edgew(\treegr) \Ccon_\treegr\big] &
\geq & \Ccon^* \sum_{\treegr \in \tract} \edgew(\treegr) \; = \;
\Ccon^* \; > \; 0\\
\sum_{\treegr \in \tract} \edgew(\treegr) \Ccon_\treegr \Ind[(s,t) \in
\edge(\treegr)] & \geq & \Ccon^* {\treegr \in \tract} \edgew(\treegr)
\Ind[(s,t) \in \edge(\treegr)] \; = \; \Ccon^* \edgew_{st} \; \geq \;
\Ccon^* \edgew^* > 0,
\end{eqnarray*}
where $\edgew^* \defn \min \limits_{(s,t) \in \edge} \edgew_{st} > 0$.
Applying these bounds to equation~\eqref{EqnTmpLower} yields the final
inequality
\begin{eqnarray}
z' \nabla^2 \Cbsur(\taupar) z & \geq & \Ccon^* \edgew^* \| z \|^2
\quad \forall z \in \real^\df
\end{eqnarray}
with $\Ccon^* \edgew^* > 0$, which establishes that the eigenvalues of
$\nabla^2 \Cbsur(\taupar)$ are bounded away from zero.

%%%%%%%%%%%%%%%%%%%%%%%%%%%%%%%%%%%%%%%%%%%%%%%%%%%%%%%%%%%%%%%%%%%%%

\section{Form of exponential parameter}
\label{AppGamForm}

Consider the observation model $y_s = \alsnr z_s + \sqrt{1-\alsnr^2}
v_s$, where $v_s \sim N(0,1)$ and $z_s$ is a mixture of two Gaussians
$(\meanvec_0, \sigma^2_0)$ and $(\meanvec_1, \sigma^2_1)$.
Conditioned on the value of the mixing indicator $X_s = j$, the
distribution of $y_s$ is Gaussian with mean $\alsnr \meanvec_j$ and
variance $\alsnr^2 \sigma^2_j + (1 - \alsnr^2)$.

Let us focus on one component $p(y_s \, | \, x_s)$ in the factorized
conditional distribution $p(y \, | \, x) = \prod_{s = 1}^n p(y_s \, |
\, x_s)$.  For $j=0, 1$, it has the form
\begin{eqnarray}
p(y_s \, | \, X_s = j) & = & \frac{1}{\sqrt{2 \pi \big[\alsnr^2
\sigma^2_j + (1 - \alsnr^2) \big] }} \exp \Big \{ - \frac{1}{2
\big[\alsnr^2 \sigma^2_j + (1 - \alsnr^2) \big] } (y_s - \alsnr
\meanvec_{j}) ^2 \Big \}.
\end{eqnarray}
We wish to represent the influence of this term on $x_s$ in the
form $\exp (\gamma_s x_s)$ for some exponential parameter $\gamma_s$.
We see that $\gamma_s$ should have the form
\begin{eqnarray*}
\gamma_s & = & \log p(y_s \, | \, X_s = 1) - \log p(y_s \, | \, X_s = 0) \\
& = & \frac{1}{2} \log \frac{\big[\alsnr^2 \sigma^2_0 + (1 - \alsnr^2)
\big]} {\big[\alsnr^2 \sigma^2_1 + (1 - \alsnr^2) \big]} + \frac{(y_s
- \alsnr \meanvec_0)^2 }{2 \big[\alsnr^2 \sigma^2_0 + (1 - \alsnr^2)
\big]} - \frac{(y_s - \alsnr \meanvec_1)^2}{2 \big[\alsnr^2 \sigma^2_1
+ (1 - \alsnr^2) \big] }
\end{eqnarray*}
 
%%%%%%%%%%%%%%%%%%%%%%%%%%%%%%%%%%%%%%%%%%%%%%%%%%%%%%%%%%%%%%%%%%%%%

%%%%%%%%%%%%%%%%%%%%%%%%%%%%%%%%%%%%%%%%%%%%%%%%%%%%%%%%%%%%%%

%%%%%%%%%%%%%%%%%%%%%%%%%%%%%%%%%%%%%%%%%%%%%%%%

\end{document}